\documentclass[10pt,twocolumn,letterpaper]{article}

\usepackage[pagenumbers]{cvpr} % To force page numbers, e.g. for an arXiv version

\usepackage{graphicx}
\usepackage{amsmath}
\usepackage{amssymb}
\usepackage{booktabs}
\usepackage{CJKutf8}
\usepackage{color}

\usepackage[pagebackref,breaklinks,colorlinks]{hyperref}

\usepackage[accsupp]{axessibility} % Improves PDF readability for those with disabilities.

\usepackage[capitalize]{cleveref}
\crefname{section}{Sec.}{Secs.}
\Crefname{section}{Section}{Sections}
\Crefname{table}{Table}{Tables}
\crefname{table}{Tab.}{Tabs.}

\def\cvprPaperID{11585} % *** Enter the CVPR Paper ID here
\def\confName{CVPR}
\def\confYear{2024}

\begin{document}
	
	\title{FaceCom: Towards High-fidelity 3D Facial Shape Completion \\
		 via Optimization and Inpainting Guidance}
	
	\author
	{
		Yinglong Li$^{1}$ , Hongyu Wu$^{1}$\thanks{Corresponding author.} , Xiaogang Wang$^{3}$ , Qingzhao Qin$^{2}$ , \\ Yijiao Zhao$^2$ , Yong Wang$^{2}$ , Aimin Hao$^{1}$ \vspace{5pt} \\
		$^1$State Key Laboratory of Virtual Reality Technology and Systems, Beihang University\\
		$^2$Peking University School and Hospital of Stomatology\\
		$^3$College of Computer and Information Science, Southwest University\\
		{\tt\small \{dragonylee, whyvrlab, ham\}@buaa.edu.cn, wangxiaogang@swu.edu.cn, }\\ 
		{\tt\small 2211210502@pku.edu.cn, \{kqcadcs, kqcadc\}@bjmu.edu.cn}
	}
\maketitle

%%%%%%%%% ABSTRACT
\begin{abstract}
We propose FaceCom, a method for 3D facial shape completion, which delivers high-fidelity results for incomplete facial inputs of arbitrary forms. Unlike end-to-end shape completion methods based on point clouds or voxels, our approach relies on a mesh-based generative network that is easy to optimize, enabling it to handle shape completion for irregular facial scans. We first train a shape generator on a mixed 3D facial dataset containing 2405 identities. Based on the incomplete facial input, we fit complete faces using an optimization approach under image inpainting guidance. The completion results are refined through a post-processing step. FaceCom demonstrates the ability to effectively and naturally complete facial scan data with varying missing regions and degrees of missing areas. Our method can be used in medical prosthetic fabrication and the registration of deficient scanning data. Our experimental results demonstrate that FaceCom achieves exceptional performance in fitting and shape completion tasks. The code is available at \href{https://github.com/dragonylee/FaceCom.git}{https://github.com/dragonylee/FaceCom.git}.
\end{abstract}

%%%%%%%% Introduction
\section{Introduction}

The shape completion of incomplete facial scans presents a significant challenge in the fields of computer vision, graphics, and medicine. On one hand, limitations in scanning devices and occlusions during the capture process often result in incomplete scans. On the other hand, clinical scenarios often involve patients with facial defects, for whom conventional maxillofacial repair methods such as mirrored reconstruction \cite{bockey2018computer} or the selection of pre-designed prosthetics from a database \cite{palousek2014use} may lack personalized shapes and require manual adjustments to fit the patient's facial contours. With the rapid development of deep learning and the increasing availability of three-dimensional facial datasets, there is a growing interest in leveraging diverse individual samples and deep learning techniques for personalized and precise facial shape completion.

While recent years have seen numerous works on image inpainting for faces \cite{lugmayr2022repaint,liu2021pd,zeng2019learning,iizuka2017globally}, there have been relatively few studies directly addressing shape completion for three-dimensional facial data. Most voxel-based shape completion approaches \cite{dai2017shape,stutz2018learning} and point cloud based methods \cite{yan2022shapeformer,xiang2021snowflakenet} have demonstrated promising results on various 3D datasets, such as ShapeNet \cite{chang2015shapenet}. However, when applied to 3D facial datasets, their completion results for incomplete data are often rough and inaccurate. These results may be attributed to the high similarity between 3D facial shapes, which requires networks to learn subtle shape variations. While point cloud-based methods can learn different shapes, they are insensitive to small-scale deformations and struggle to generate smooth results due to the lack of explicit surface adjacency. Similarly, voxel based methods, although offering regular results and ease of construction and training, face challenges in generating high-precision results due to memory and resolution limitations. Therefore, we propose a novel approach that operates directly on facial mesh data for generation purposes.

Our proposed method involves a graph convolutional neural network structure, specifically an autoencoder, that operates directly on explicit meshes to learn various facial shape features and perform parameterized generation tasks. Our approach aims to learn the facial shapes of as many identities as possible, focusing on neutral faces without considering variations of expressions. To address the limited number of identities in existing 3D facial datasets, we mix several publicly available datasets with our collected facial dataset, resulting in a comprehensive 3D facial dataset with a substantial number of identities and unified topology. Following auto-regressive training, we obtain a shape generator capable of sampling from a hypersphere distribution and generating a complete facial mesh. Additionally, leveraging the larger scale of facial image datasets, we train a 2D facial inpainting network applicable to our dataset based on the work by Lugmayr $\etal$ \cite{lugmayr2022repaint}. In the case of incomplete facial inputs, we utilize an optimization method to fit a complete facial mesh, with guidance provided by the results of image inpainting. Post-processing is employed to refine and achieve the final completion results. Please refer to \cref{fig:shape-completion} for the visual results.

Our approach excels in generating high-fidelity completion results for various forms of incomplete input, without the need for specific vertex numbers or correspondences. In our experiments, we demonstrate that our method outperforms other state-of-the-art models in fitting neutral faces and exhibits excellent completion results for various incomplete faces. Furthermore, our method yields practical application results in clinical data completion experiments.

In summary, the main contributions of this paper include:
\begin{itemize}
	\setlength{\itemsep}{0pt}
	\setlength{\parsep}{0pt}
	\setlength{\parskip}{0pt}
	\item Introducing a novel 3D facial completion method based on generative model, image inpainting guidance and optimization techniques, capable of generating high-fidelity completion results for free-form incomplete facial scans.
	\item Proposing a three-dimensional face generation network based on a multi-scale encoder-decoder architecture, outperforming existing state-of-the-art models in fitting neutral faces.
	\item Validating the potential applications of our completion method in clinical maxillofacial repair and non-rigid registration fields.
\end{itemize}

%%%%%%%%% Related Work
\section{Related Work}

\textbf{Parametric Face and Head Models}. Blanz and Vetter\cite{blanz2023morphable} were the pioneers in introducing 3D Morphable Models (3DMMs) for representing faces. They utilized PCA to reduce the dimensions of shape and texture spaces, enabling parametric face generation. Subsequently, 3DMMs found widespread application in face reconstruction\cite{feng2021learning,wood20223d,zollhofer2018state}. Based on our research, current studies on parameterized face or head models can be broadly categorized into three types. The first type involves linear or blendshapes-based statistical models\cite{paysan20093d,brunton2014multilinear,cao2013facewarehouse,bolkart2015groupwise,booth2018large,dai20173d,dai2020statistical,li2017learning,yang2020facescape}. These models often use a unified topology dataset and separately model shape, expression, or texture, and are characterized by their ease of construction and processing. The second type aims to achieve non-linear deformation explicitly using deep learning methods\cite{ranjan2018generating,bouritsas2019neural,chen2021learning,cheng2019meshgan}, which extract features with spectral or spiral graph convolutions. The final type represents the recent emergence of implicit neural functions for constructing continuous models\cite{zheng2022imface,wang2022morf,yenamandra2021i3dmm,giebenhain2023learning}. These models can finely reproduce facial expression variations but often exhibit poorer fitting performance for different identities.

\textbf{Shape Completion}. 3D shape completion, or point cloud completion, is a technique used to reconstruct missing parts of input models. In recent years, deep learning methods have shown promise in effectively addressing missing data in 3D models. Some approaches have focused on learning and generating based on voxels due to their regularity and ease of convolutional processing\cite{dai2017shape,stutz2018learning,wang2021voxel}. However, voxel-based methods tend to consume a significant amount of space and yield lower-quality results. Consequently, several studies have directly tackled shape completion tasks on point cloud\cite{tchapmi2019topnet,xiang2021snowflakenet,yu2021pointr} or have combined graph convolutions\cite{litany2018deformable,zhu2021towards} to incorporate more local information. Recently some novel methods have also leveraged implicit functions to avoid discretization or have integrated new generative models to achieve superior generation outcomes\cite{yan2022shapeformer,cheng2023sdfusion}. Despite these advancements, the effectiveness of these shape completion techniques on highly detailed facial data remains limited. The work of Litany \etal\cite{litany2018deformable} shares some similarities with our approach and also attempted facial completion experiments, yet they imposed constraints on the topological structure of incomplete input, and the completion results were found to be unsatisfactory.

\begin{figure}[t]
	\centering
	\includegraphics[width=1\linewidth]{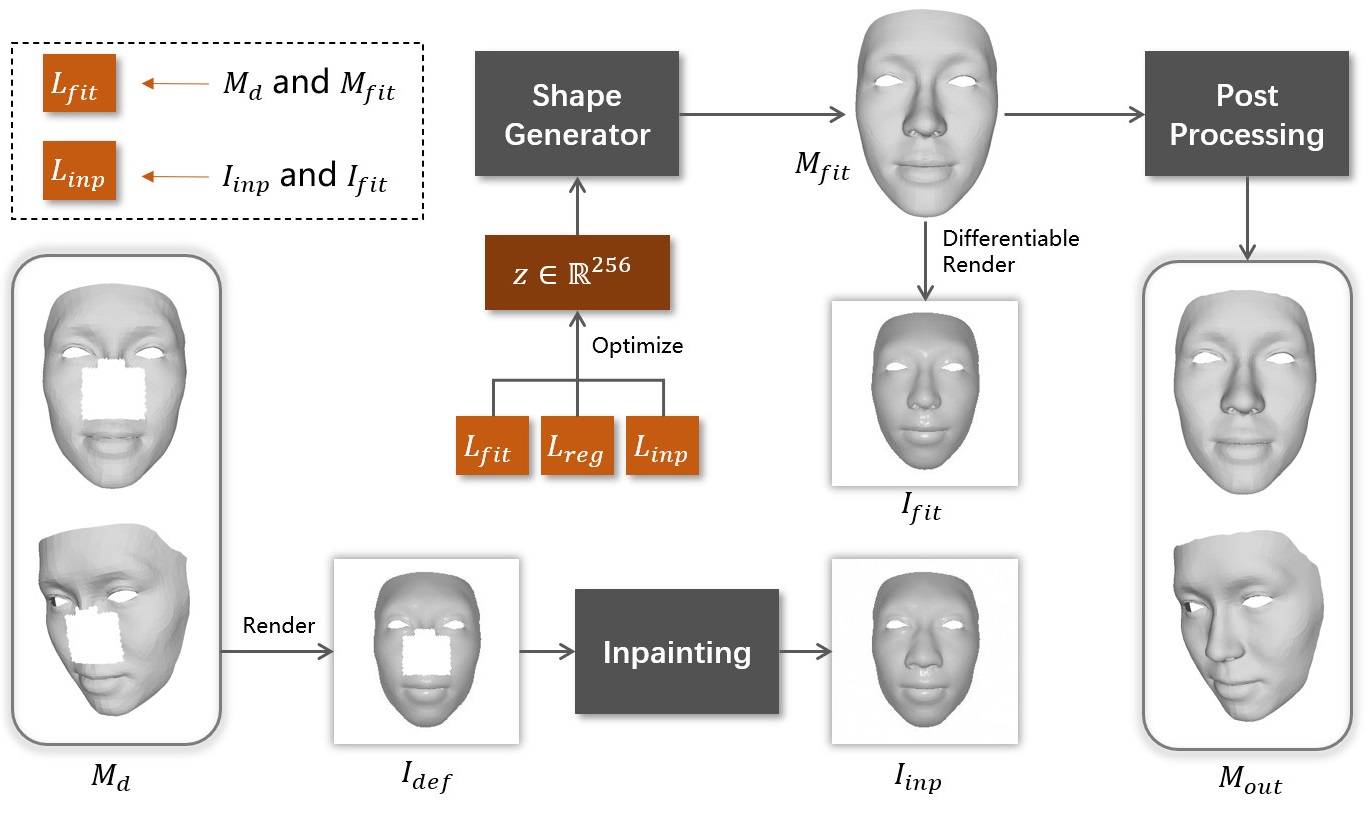}
	\caption{\textbf{FaceCom Overview}. During the shape fitting stage, the shape generator is employed to fit the incomplete facial scan $M_{d}$ using an optimization method. The optimization objective comprises several loss functions, including the disparity between the input $M_{d}$ and the fitting result $M_{fit}$, the difference between the images $I_{fit}$ and $I_{inp}$, and regularization terms. Following the acquisition of the optimal complete shape relative to the incomplete input, a post-processing step is applied to achieve the final completion results $M_{out}$.}
	\label{fig:method}
\end{figure}

%%%%%%%%%%% Method
\section{Method}
\label{method}

We propose an algorithm called FaceCom, which is designed for shape completion of incomplete 3D facial scans. The key components of our completion process, including the shape generator, image inpainting guidance, and post-processing, will be discussed in this section. The workflow of FaceCom is illustrated in \cref{fig:method}.

\subsection{Shape Generator}
\label{shape-generator}

\begin{figure*}
	\centering
	\includegraphics[width=0.94\linewidth]{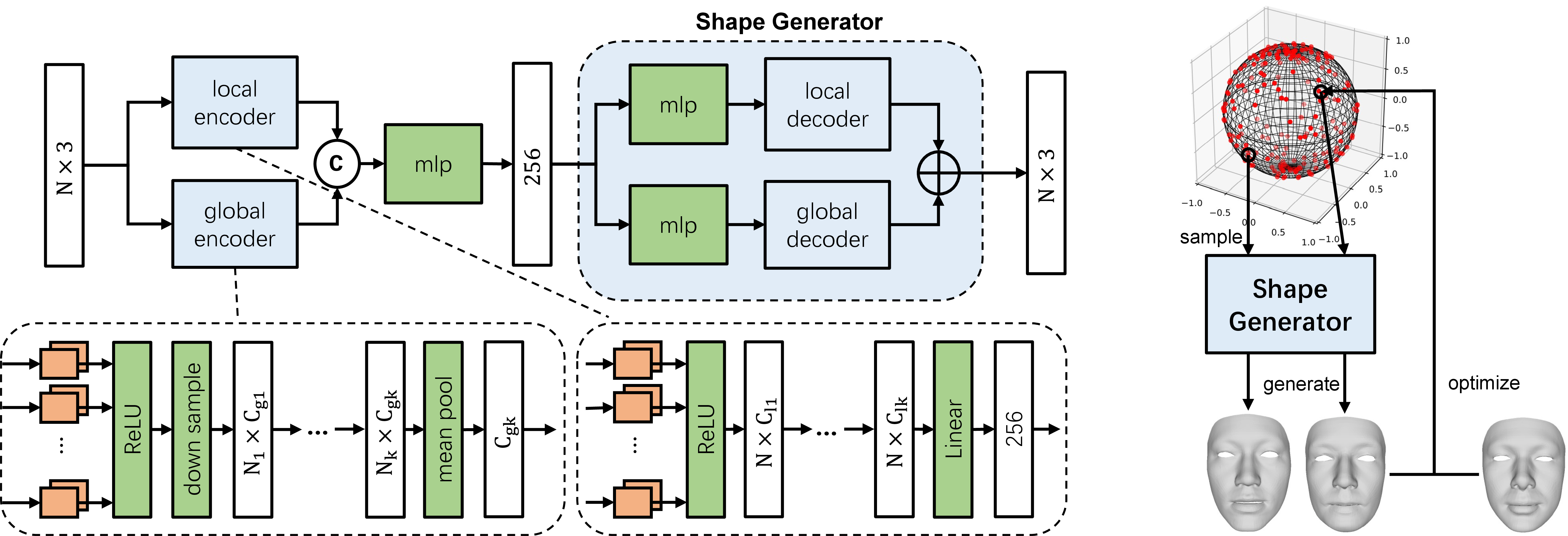}
	\caption{\textbf{Architecture of shape generator}. Left: Network used for training, which is primarily based on a graph convolutional autoencoder structure. It conducts feature extraction and generation separately from both local and global perspectives. After training, the decoder component serves as the shape generator for FaceCom. Right: Our shape generator can produce diverse facial samples from the hypersphere space and is readily optimized within the latent space.}
	\label{fig:network}
\end{figure*}

We employ a GNN-based autoencoder framework for learning the representation of 3D facial meshes and generation, as shown in \cref{fig:network}. The procedure involves extracting features from the input facial mesh at both the global and local levels, with the fused features producing a 256-dimensional implicit representation of the face. The shape generation phase worked inversely to the encoding phase, decoding the implicit vector to gain a complete facial shape. The encoder part is meticulously illustrated in \cref{fig:network}, and the decoder follows a symmetric structure. In contrast to the global path, the local path removes the downsampling layers, reduces the number of channels, and preserves local geometric information by using linear layers in place of pooling layers.

Meshes retain rich local geometric information, yet their irregularity makes them challenging to process. We employed FeaStNet\cite{verma2018feastnet, Fey/Lenssen/2019} to conduct convolution operations directly on meshes, which demonstrates excellent feature extraction capabilities. Downsampling layers were implemented and improved based on the work of Ranjan \etal\cite{ranjan2018generating}. To ensure the differentiability of the downsampling process, we simplified\cite{garland1997surface} the face template $M$, resulting in $M_1,\dots,M_k$, where $M_i\in\mathbb{R}^{N_i\times 3}$. We assumed that a vertex's position in $M_{i+1}$ is a linear combination of the positions of the k-nearest vertices in $M_i$, allowing us to compute the transformation matrix $M_{i+1}=D_i M_i$, where the sparse matrix $D_i\in\mathbb{R}^{N_{i+1}\times N_i}$. Upsampling can also be achieved using a similar method. Within the neural network, we maintained fixed transformation matrices to facilitate differentiable upsampling and downsampling operations.

Our objective is to perform shape completion on incomplete(defect) facial scans in a clinical setting. Hence, we disregard facial expression variations, allowing the shape generator to learn as many individual facial shapes as possible. Unlike 2D images, publicly available 3D facial datasets often contain a limited number of individuals. To obtain a more diverse and aligned dataset, we applied several public 3D facial datasets\cite{dai2020statistical,yang2020facescape} to use their neutral faces. Additionally, we collected 400 neutral facial scans in the hospital. All the 3D facial data we utilized is presented in \cref{tab:dataset}. Subsequently, we employed the non-rigid registration method proposed in \cite{fan2023towards} to achieve a dataset $\mathcal{X}$ with a unified topological structure, which comprises numerous individuals, different regions, and various age groups. Around 5\% of the data are reserved for subsequent testing purposes.

\begin{table}
	\centering
	%\begin{tabular}{@{}lc@{}l@{}}
	\begin{tabular}{lcc}
		\toprule
		Dataset & Sub. Num. & Reserved \\
		\midrule
		Headspace\cite{dai2020statistical,zielonka2022towards,dai20173d} & 1171 & 57\\
		FaceScape\cite{yang2020facescape} & 834 & 41\\
		Ours & 400 & 20\\
		\midrule
		Total & 2405 & 118 \\
		\bottomrule
	\end{tabular}
	\caption{\textbf{3D face datasets used in our study}, all aligned to a unified topology with approximately 5\% reserved for testing. Note that low-quality samples were manually excluded, resulting in a reduced number of subjects from the origins.}
	\label{tab:dataset}
\end{table}

During training, the reconstruction loss of the network is defined as the mean squared error between the input $\mathbf{X}$ and the output $\mathit{dec}(\mathit{enc}(\mathbf{X}))$, denoted as $L_{MSE}$. Since the autoencoder essentially serves as a data compression method, facilitating a point-to-point mapping between data samples and latent space points, its inherent nature lacks generative capabilities. As a result, we attempted to train the model using the Variational Autoencoder (VAE) approach \cite{kingma2013auto}. However, we encountered challenges such as the vanishing KL divergence and non-decreasing $L_{MSE}$, hindering the network from reaching a satisfactory convergence. Consequently, we introduced a regularization constraint expressed by
\begin{equation}
	L_{reg}=(||\mathbf{z}||_2 -1)^2,
	\label{eq:regularization}
\end{equation}
aiming to confine the mesh's implicit representation, denoted as $\mathbf{z}=\mathit{enc}(\mathbf{X})$, within a hypersphere distribution. Our experimental results further supported the notion that this approach facilitates optimization-based inference procedures. Prior study \cite{zhao2018adversarially} has illustrated that this approach can significantly enhance the autoencoder's generative capacity and overall stability. The optimization objective was defined as $L_{MSE}+\lambda L_{reg}$. We initialized the learning rate at 0.001 and halved it every 50 epochs. The training process utilized the Adam optimizer \cite{kingma2014adam} within the PyTorch framework, with the entire training process taking approximately 6 hours on an NVIDIA Geforce RTX4090 GPU.

\subsection{Shape Fitting}
\label{completion-by-optimization}

Many generation-based image inpainting approaches \cite{iizuka2017globally,liu2021pd} and point cloud shape completion techniques \cite{dai2017shape,zhou20213d} train an end-to-end network to directly complete the missing parts of the input. However, incomplete meshes lose topological information, making it challenging to seamlessly grow the completion parts using deep learning methods. Hence, we first trained a parameterized generative model (refer to \cref{shape-generator}), followed by using an optimization-based method to fit the input of the incomplete facial mesh.

Let $\mathcal{D}(\mathbf{v},\mathbf{M})$ denote the distance from vertex $\mathbf{v}$ to the surface $\mathbf{M}$. For surfaces represented by triangle meshes, we calculate the closest distance from points to the triangular faces. For the incomplete facial mesh input $\mathbf{M}_d$, the fitting loss is defined as
\begin{equation}
	L_{fit}=\frac{1}{|\mathbf{M}_d|}\sum_{\mathbf{v}\in\mathbf{M}_d}\mathcal{D}(\mathbf{v},\mathcal{T}(\mathit{dec}(\mathbf{z}))),
	\label{eq:fit}
\end{equation}
which quantifies the similarity between generated mesh and the input incomplete mesh, where $\mathcal{T}$ represents a rigid transformation. In practice, the incomplete facial mesh may exhibit inconsistencies relative to the template, so we typically exclude a certain percentage of vertices farthest from the calculation of $L_{fit}$. Drawing upon the above procedures, the shape fitting process can be described as
\begin{equation}
	\min_{\mathbf{z},\mathcal{T}} \ L_{fit}+\lambda L_{reg}.
	\label{eq:completion}
\end{equation}

During the fitting process, we implemented a differentiable $\mathcal{D}$ based on publicly available PyTorch3D \cite{ravi2020pytorch3d}, obtaining the optimal fitting result $\mathit{dec}(\mathbf{z})$ through iterative optimization. Notably, our optimization-based method is applicable to any form of incomplete facial input, allowing $\mathbf{M}_d$ to be a mesh, a point cloud, or a set of key points.

\subsection{Image Inpainting Guidance}
\label{image-inpainting-guidance}

Our facial shape generator was trained on a dataset including neutral faces from thousands of identities. However, this volume is relatively small compared to the datasets used in the field of image generation \cite{dhariwal2021diffusion,lugmayr2022repaint}, making it difficult to cover a wide range of practical scenarios. Therefore, a natural idea is to train an image inpainting network customized for our dataset, which can effectively guide the three-dimensional shape completion process.

We utilize RePaint\cite{lugmayr2022repaint}, essentially a denoising diffusion probabilistic model\cite{ho2020denoising,dhariwal2021diffusion}, as a prior for our image inpainting process. The intermediate image $x_t$ in the reverse diffusion process originates from $x_{t+1}$ and the conditional probability $p_{\theta}(x_t|x_{t+1})$. RePaint introducing a mask $m$ as input, known parts $m(x_0)$ of the image are sampled during the diffusion process using the conditional probability $q(x_t|x_0)$, resulting $m(x_t^{\text{known}})$, while the unknown parts retain the results of the inverse diffusion process $(1-m)(x_t^{\text{unknown}})$. The combined result serves as the intermediate sampling result, effectively incorporating guided information into the image generation process. Additionally, a resampling technique is employed to improve the coherence between the repaired and known parts. Further details can be found in \cite{lugmayr2022repaint}.

An image inpainting network tailored to our specific task, denoted as $\mathit{inp}$, was trained to guide the three-dimensional shape completion process. We first set up a differentiable rendering camera $\mathcal{R}$ with fixed parameters. Subsequent, leveraging the network weights from CelebA-HQ \cite{liu2015deep} in \cite{lugmayr2022repaint} and the training approach outlined in \cite{dhariwal2021diffusion}, we fine-tuned the network on our dataset $\mathcal{R}(\mathcal{X})$, resulting in a network capable of taking incomplete rendered faces and masks as input and producing restoration outputs.

Define the loss function between the image inpainting results and the rendering results of the fitting shape as
\begin{equation}
	L_{inp}=\mathit{MSE}( \mathcal{R}(\mathcal{T}(\mathit{dec}(\mathbf{z}))), \mathit{inp}(\mathcal{R}(\mathbf{M}_d))).
	\label{eq:inpaint}
\end{equation}
By integrating the image inpainting guidance, the shape completion process for incomplete facial scans can be described as
\begin{equation}
	\min_{\mathbf{z},\mathcal{T}} \ L_{fit}+\lambda_1 L_{inp}+\lambda_2 L_{reg}.
	\label{eq:completion_guide}
\end{equation}

\begin{figure*}
	\centering
	\begin{subfigure}{0.13\linewidth}
		\includegraphics[width=0.8\linewidth]{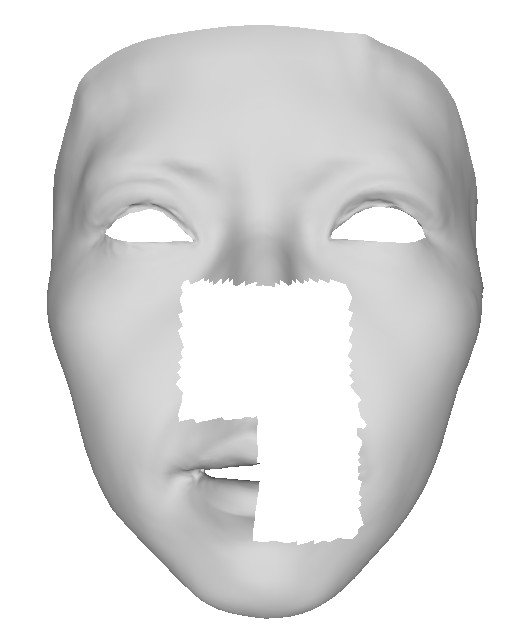}
		\caption{defect input}
		\label{fig:post-a}
	\end{subfigure}
	\centering
	\begin{subfigure}{0.13\linewidth}
		\includegraphics[width=0.8\linewidth]{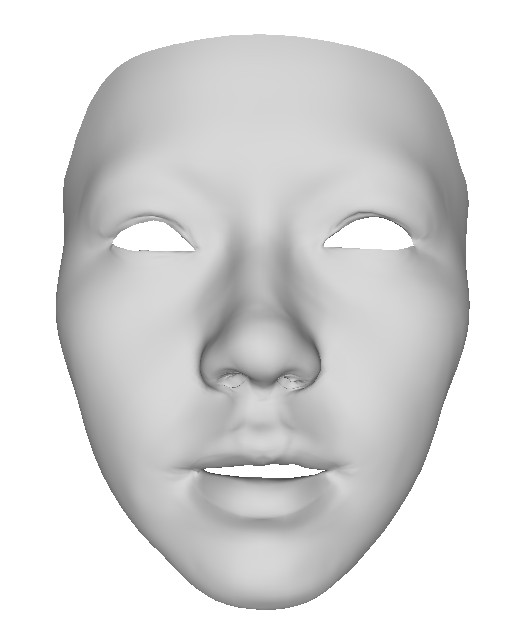}
		\caption{fit}
		\label{fig:post-b}
	\end{subfigure}
	\centering
	\begin{subfigure}{0.13\linewidth}
		\includegraphics[width=0.8\linewidth]{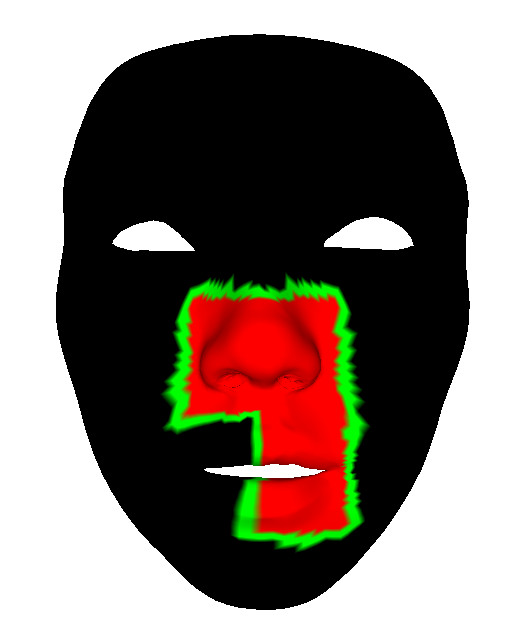}
		\caption{identification}
		\label{fig:post-c}
	\end{subfigure}
	\centering
	\begin{subfigure}{0.13\linewidth}
		\includegraphics[width=0.8\linewidth]{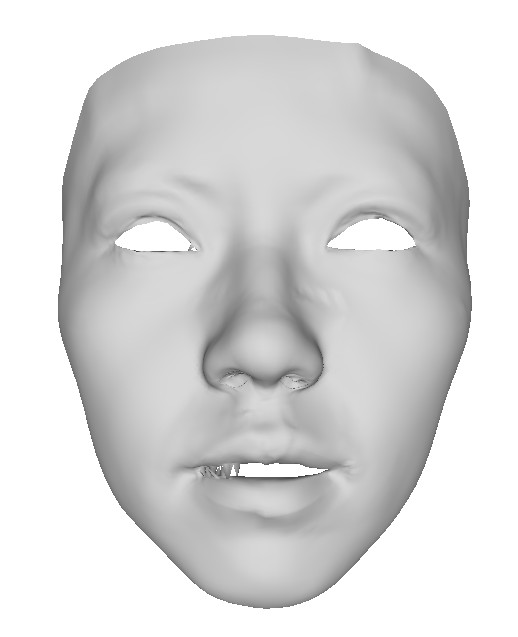}
		\caption{projection}
		\label{fig:post-d}
	\end{subfigure}
	\centering
	\begin{subfigure}{0.13\linewidth}
		\includegraphics[width=0.8\linewidth]{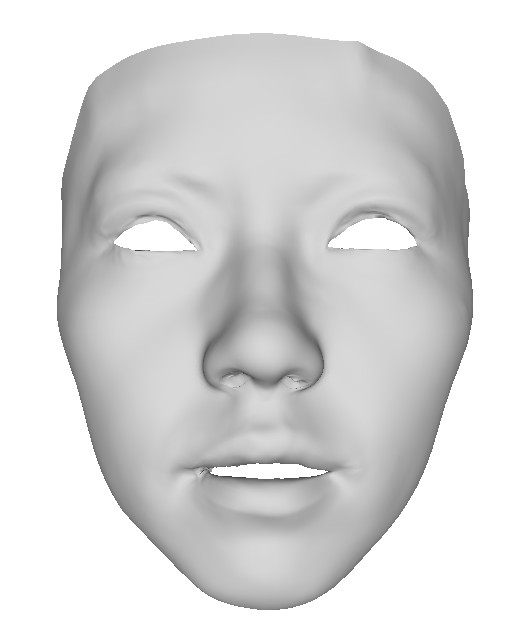}
		\caption{refinement}
		\label{fig:post-e}
	\end{subfigure}
	\centering
	\begin{subfigure}{0.13\linewidth}
		\includegraphics[width=0.8\linewidth]{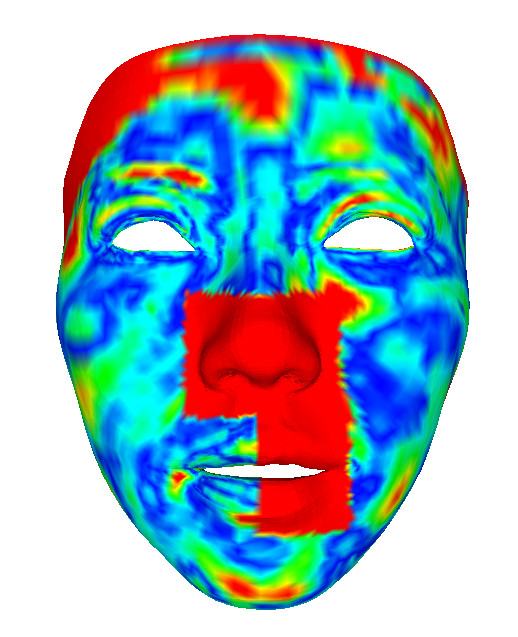}
		\caption{\subref{fig:post-a} vs. \subref{fig:post-b}}
		\label{fig:post-f}
	\end{subfigure}
	\centering
	\begin{subfigure}{0.13\linewidth}
		\includegraphics[width=0.8\linewidth]{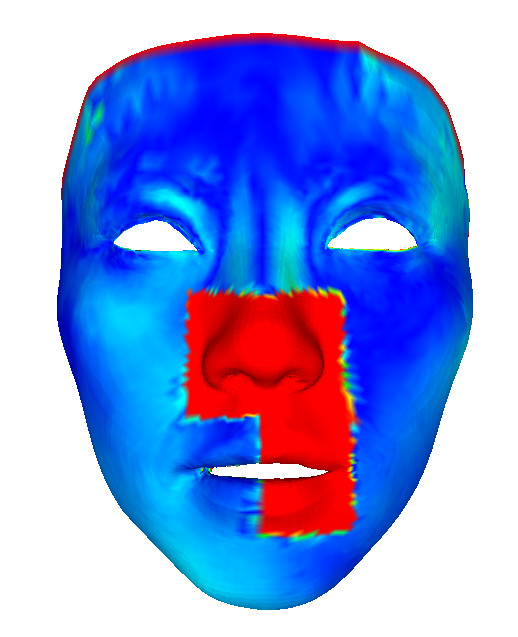}
		\caption{\subref{fig:post-a} vs. \subref{fig:post-e}}
		\label{fig:post-g}
	\end{subfigure}
	\centering
	\begin{subfigure}{0.03\linewidth}
		\includegraphics[width=1\linewidth]{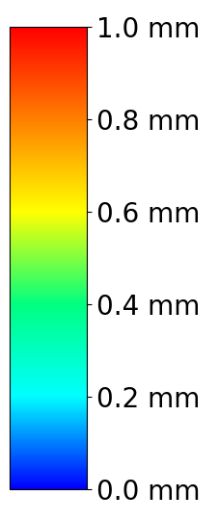}
		\label{fig:post-h}
	\end{subfigure}
	
	\caption{\textbf{Visualization of post-processing technique}. \subref{fig:post-a} depicts the incomplete facial scan input, \subref{fig:post-b} demonstrates the fitting result, \subref{fig:post-c} to \subref{fig:post-e} illustrate each step of the post-processing process. \subref{fig:post-f} and \subref{fig:post-g} showcase the disparities between the results before and after post-processing and the incomplete input.}
	\label{fig:post}
\end{figure*}

\subsection{Post Processing}
\label{post-processing}

Our optimization-based approach is capable of completing various types of incomplete 3D facial input. However, from an application standpoint, inconsistencies in the non-defective regions of the completion results can lead to misalignments at the boundaries. To address this limitation, we have devised a series of engineering post-processing methods. Through post-processing, the completed results exhibit greater consistency with the defective input, leading to more precise completion outcomes.

The visualization of the post-processing is illustrated in \cref{fig:post}. For the defective facial input \cref{fig:post-a}, we employed the techniques described in Sections \cref{shape-generator} and \cref{image-inpainting-guidance} to obtain the fitted face \cref{fig:post-b}. Subsequently, we identified the repaired parts of \cref{fig:post-b} using a thresholding approach, as shown in \cref{fig:post-c}. We then projected the non-defective regions onto the surface represented by the defect input along the vertex normal direction, yielding the result displayed in \cref{fig:post-d}. Finally, by leveraging the K-nearest neighbors of the extended portion of the defective region (the green portion in Figure \cref{fig:post-c}), we performed weighted vertex deformation to obtain the final completion outcome \cref{fig:post-e}. The projection process may introduce some outliers, therefore there is an additional refinement step in the end.

Please note that the post-processing stage primarily serves as an engineering approach to enhance the conformity between the completion results and the defect input. It addresses inherent issues within the optimization-based method. Post-processing greatly refines the final results particularly for faces with pronounced wrinkles. While we provide a brief description of the post-processing stage here, more detailed information is available in our supplementary materials. \cref{fig:post-f,fig:post-g} demonstrate the disparities of the fitted face and the final completion result, relative to the defective facial input, using the same scale for color mapping. It is evident that post-processing has yielded improved results.

\section{Experiments}
\label{experiments}

In this section, we conduct multiple experiments to test FaceCom thoroughly. We evaluate the fitting capability of the shape generator on neutral faces in \Cref{fitting}, examine FaceCom's proficiency in completing irregular incomplete facial inputs in \Cref{shape-completion}, assess FaceCom's effectiveness in designing prosthetics for clinical facial defect cases in \Cref{clinical-experiment}, explore the potential of FaceCom for non-rigid registration in \Cref{non-rigid-registration}, and finally validate the necessity of different components of FaceCom in \Cref{ablation-study}.

\subsection{Fitting}
\label{fitting}

\begin{figure*}
	\centering
	\begin{subfigure}{0.96\linewidth}
		\includegraphics[width=1\linewidth]{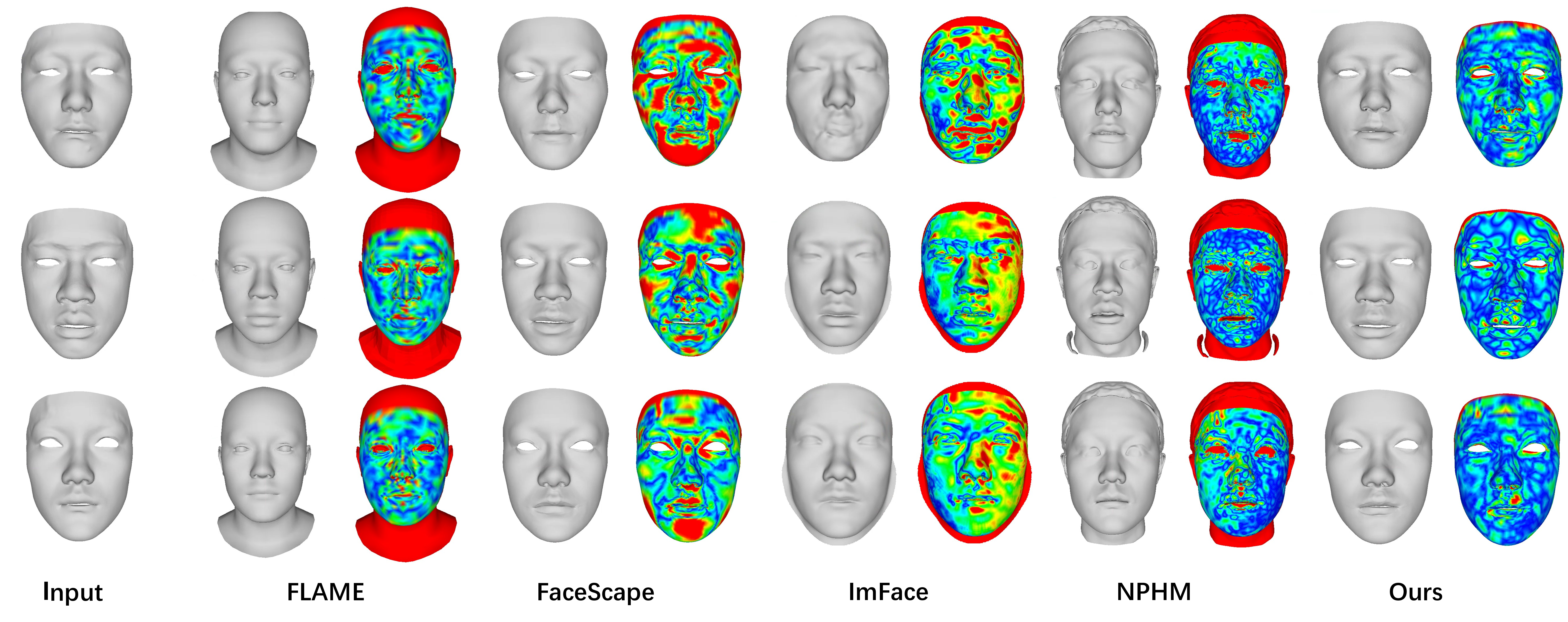}
	\end{subfigure}
	\centering
	\begin{subfigure}{0.03\linewidth}
		\includegraphics[width=1\linewidth]{figures/colorbar}
		\vspace{10pt}
	\end{subfigure}
	
	\caption{\textbf{Fitting experiment examples.}}
	\label{fig:fit}
\end{figure*}

The reserved data from \cref{tab:dataset} are resampled to create a facial scan test set, with the number of triangular faces in each mesh randomly ranging between 10k and 20k. We conducted fitting experiments on the test set using the shape generator trained in \cref{shape-generator} and compared the results with state-of-the-art 3DMM models FLAME \cite{li2017learning} and FaceScape \cite{yang2020facescape}, and the implicit parametric model ImFace \cite{zheng2022imface} and NPHM \cite{giebenhain2023learning}. Some visual examples are shown in \cref{fig:fit}.

FLAME\cite{li2017learning} and ImFace \cite{zheng2022imface} offer dedicated methods for fitting scanned data, requiring the provision of additional landmarks. We prepared compliant landmarks beforehand on the test set to facilitate the evaluation process. FaceScape \cite{yang2020facescape}, on the other hand, solely provides a method for fitting 3D key points, and NPHM \cite{giebenhain2023learning} only provides a point cloud fitting method after depth observation for specific samples. We modified their code to achieve the fitting of scanned data after spatial coordinate alignment. Our evaluation included the assessment of the L2 Chamfer Distance and the average point-to-surface distance, both are unidirectional as the fitted mesh typically has a larger scope. The Chamfer Distance(CD) is significantly influenced by vertex density, so we employed Loop Subdivision \cite{loop1987smooth} to refine the fitted meshes to approximately 200k faces before calculation. The mean point-to-surface distance(MD) corresponds to the calculation of the average $\mathcal{D}$ mentioned in \cref{completion-by-optimization}. The results, outlined in \cref{tab:fit-compare}, demonstrate the superior performance of our shape generator in fitting neutral faces compared to existing state-of-the-art 3D facial generation models.

\begin{table}
	\centering
	\begin{tabular}{lcc}
		\toprule
		Method & CD(mm)$\downarrow$ &  MD(mm)$\downarrow$\\
		\midrule
		FLAME\cite{li2017learning}& 0.622 & 0.309\\
		facescape\cite{yang2020facescape}  & 0.685 & 0.416 \\
		ImFace\cite{zheng2022imface} &  0.632 & 0.505 \\
		NPHM\cite{giebenhain2023learning} & 0.578 & 0.381 \\
		ours(w/o post-processing)  & \textbf{0.480} & \textbf{0.275} \\
		\midrule
		ours(w/ post-processing)  & & 0.160\\
		\bottomrule
	\end{tabular}
	\caption{\textbf{Fitting comparison results} on L2 chamfer distance and mean point-to-surface distance. The bottom row shows FaceCom's outcomes for non-rigid face registration.}
	\label{tab:fit-compare}
\end{table}

\subsection{Shape Completion}
\label{shape-completion}

\begin{figure*}
	
	% 1-8
	
	\centering
	\begin{minipage}{0.02\linewidth}
		\rotatebox{90}{\textbf{defect}}
	\end{minipage}
	\begin{minipage}{0.93\linewidth}
		\centering
		\begin{subfigure}{0.12\linewidth}
			\includegraphics[width=0.9\linewidth]{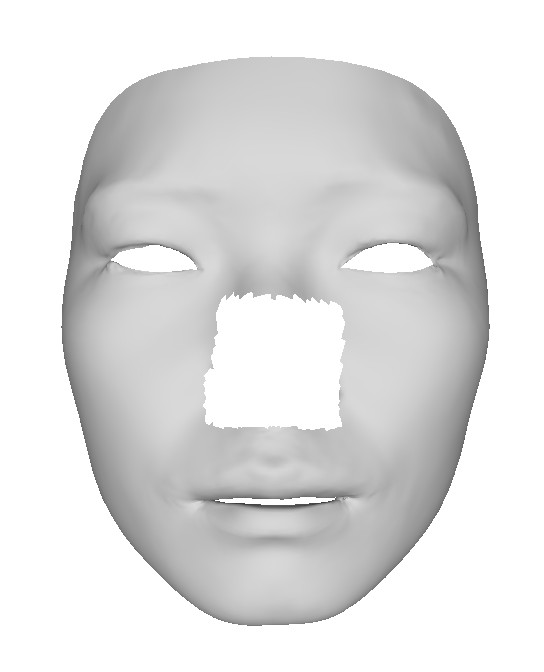}
		\end{subfigure}
		\centering
		\begin{subfigure}{0.12\linewidth}
			\includegraphics[width=0.9\linewidth]{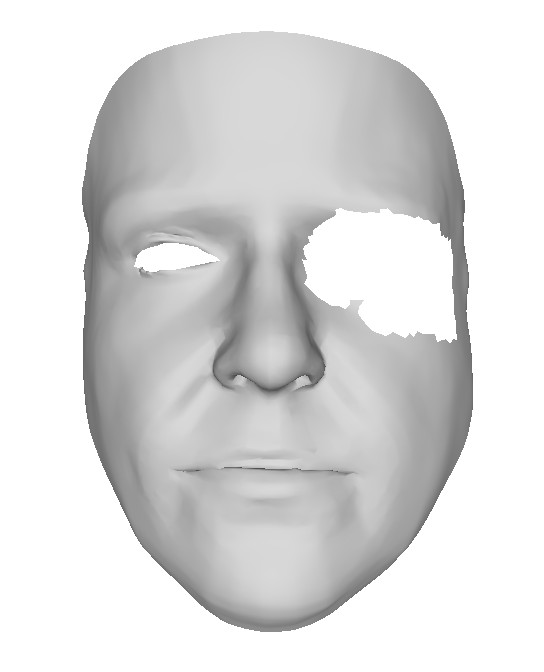}
		\end{subfigure}
		\centering
		\begin{subfigure}{0.12\linewidth}
			\includegraphics[width=0.9\linewidth]{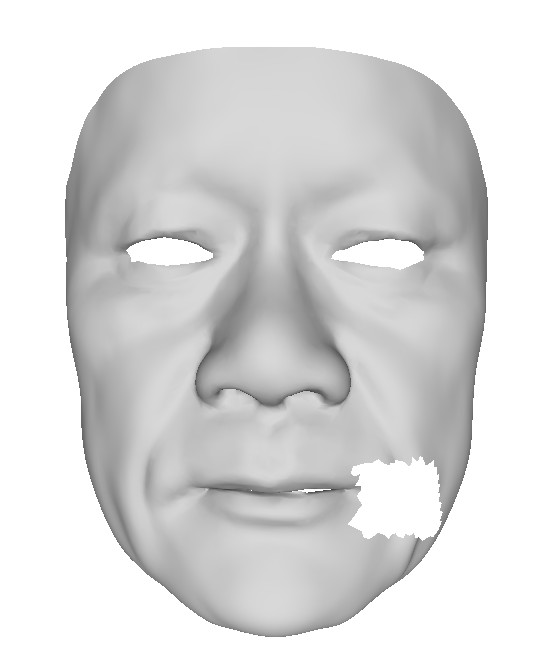}
		\end{subfigure}
		\centering
		\begin{subfigure}{0.12\linewidth}
			\includegraphics[width=0.9\linewidth]{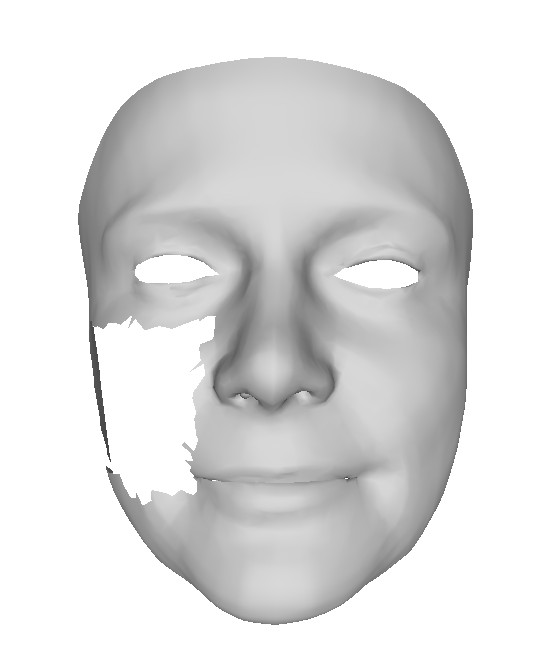}
		\end{subfigure}
		\centering
		\begin{subfigure}{0.12\linewidth}
			\includegraphics[width=0.9\linewidth]{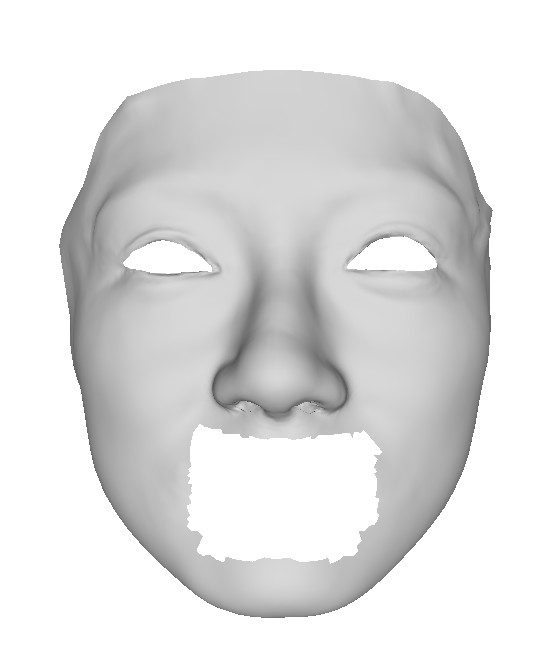}
		\end{subfigure}
		\centering
		\begin{subfigure}{0.12\linewidth}
			\includegraphics[width=0.9\linewidth]{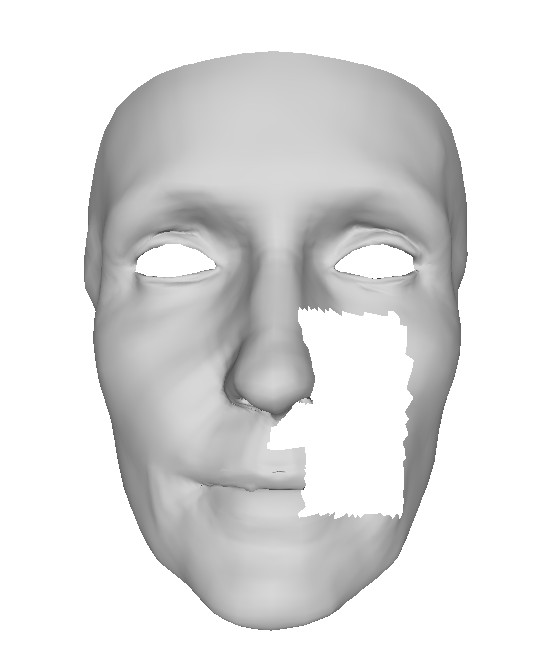}
		\end{subfigure}
		\centering
		\begin{subfigure}{0.12\linewidth}
			\includegraphics[width=0.9\linewidth]{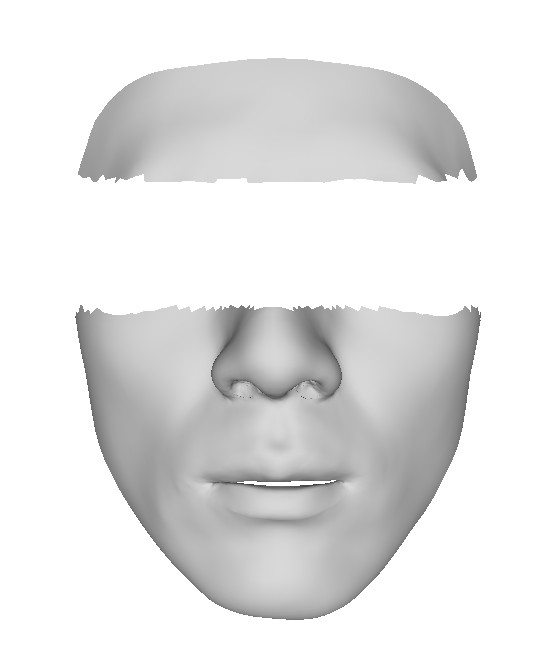}
		\end{subfigure}
		\centering
		\begin{subfigure}{0.12\linewidth}
			\includegraphics[width=0.9\linewidth]{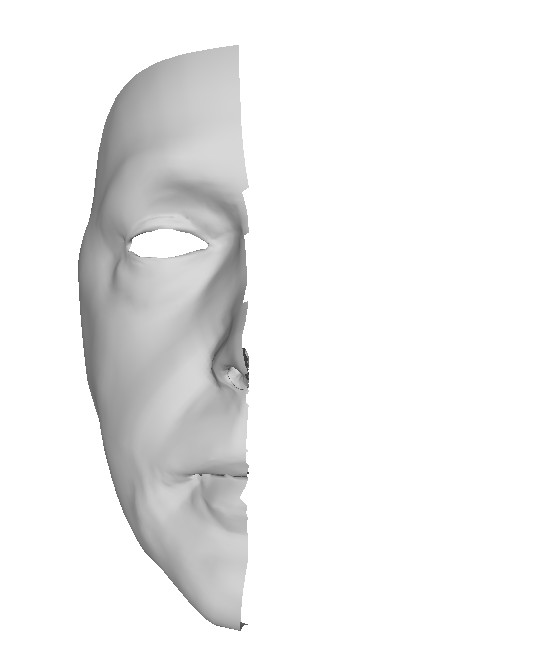}
		\end{subfigure}
	\end{minipage}
	
	\centering
	\begin{minipage}{0.02\linewidth}
		\rotatebox{90}{\textbf{result}}
	\end{minipage}
	\begin{minipage}{0.93\linewidth}
		\centering
		\begin{subfigure}{0.12\linewidth}
			\includegraphics[width=0.9\linewidth]{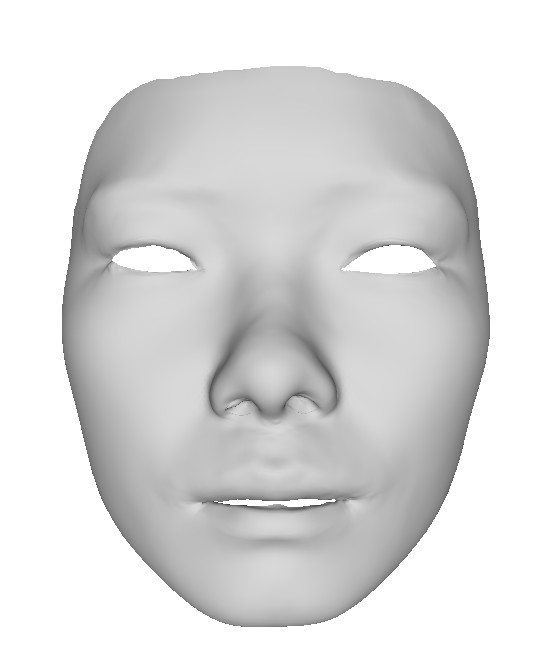}
		\end{subfigure}
		\centering
		\begin{subfigure}{0.12\linewidth}
			\includegraphics[width=0.9\linewidth]{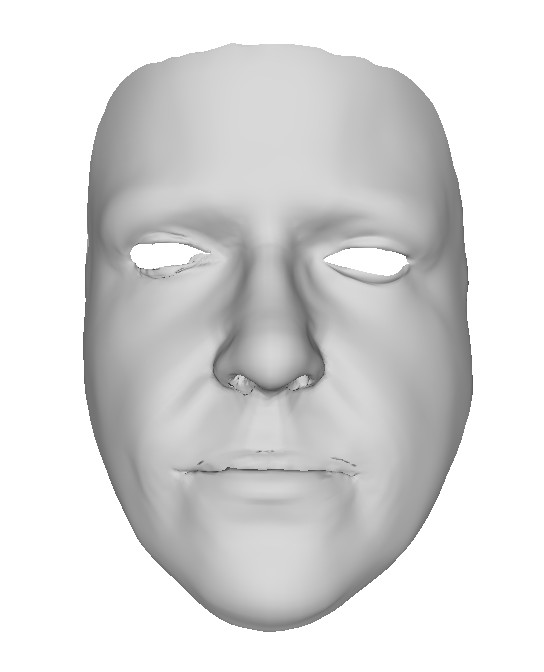}
		\end{subfigure}
		\centering
		\begin{subfigure}{0.12\linewidth}
			\includegraphics[width=0.9\linewidth]{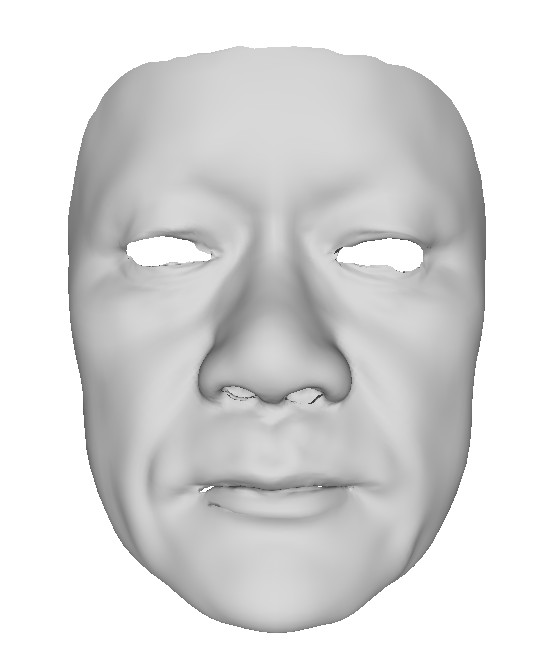}
		\end{subfigure}
		\centering
		\begin{subfigure}{0.12\linewidth}
			\includegraphics[width=0.9\linewidth]{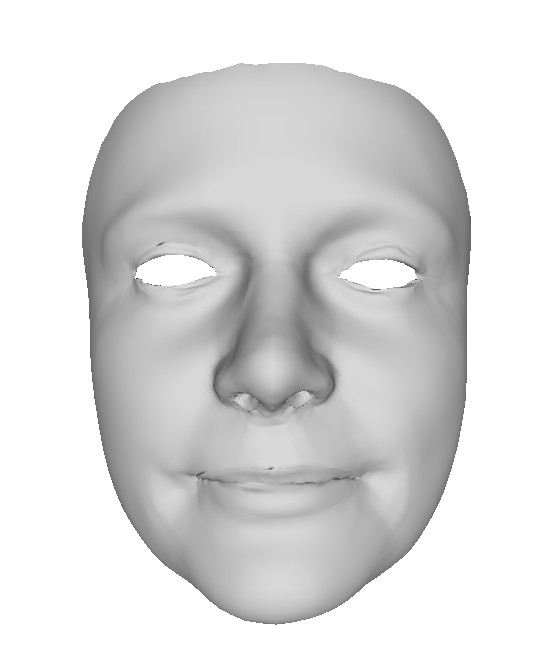}
		\end{subfigure}
		\centering
		\begin{subfigure}{0.12\linewidth}
			\includegraphics[width=0.9\linewidth]{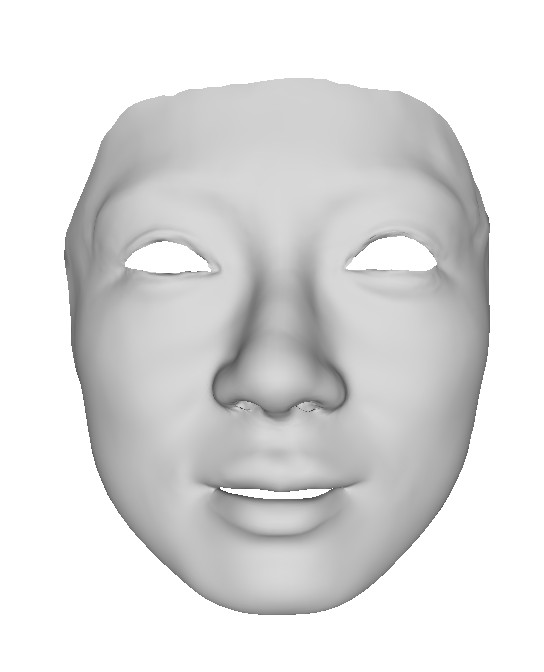}
		\end{subfigure}
		\centering
		\begin{subfigure}{0.12\linewidth}
			\includegraphics[width=0.9\linewidth]{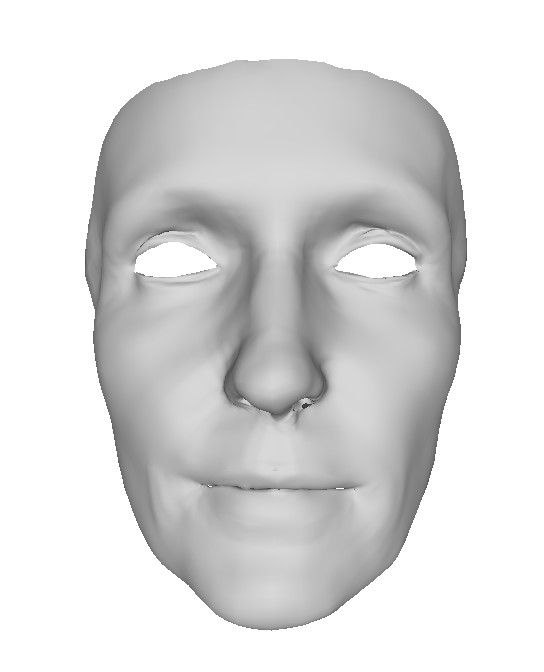}
		\end{subfigure}
		\centering
		\begin{subfigure}{0.12\linewidth}
			\includegraphics[width=0.9\linewidth]{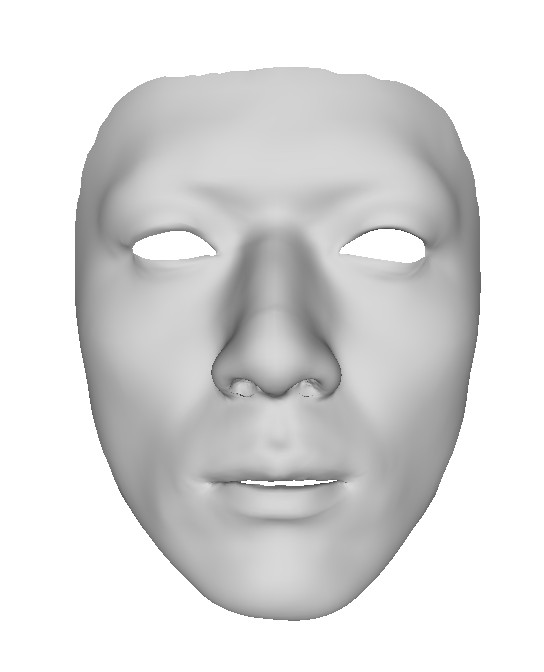}
		\end{subfigure}
		\centering
		\begin{subfigure}{0.12\linewidth}
			\includegraphics[width=0.9\linewidth]{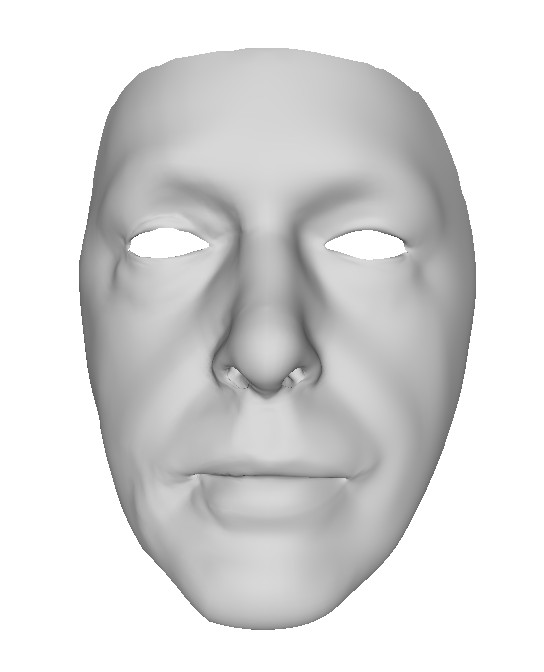}
		\end{subfigure}
	\end{minipage}
	
	\centering
	\begin{minipage}{0.02\linewidth}
		\rotatebox{90}{\textbf{gt}}
	\end{minipage}
	\begin{minipage}{0.93\linewidth}
		\centering
		\begin{subfigure}{0.12\linewidth}
			\includegraphics[width=0.9\linewidth]{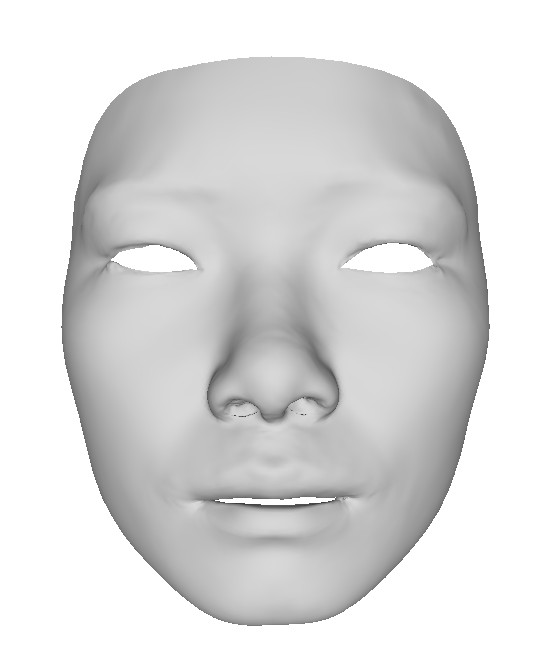}
			\subcaption{}
			\label{fig:shape-completion-a}
		\end{subfigure}
		\centering
		\begin{subfigure}{0.12\linewidth}
			\includegraphics[width=0.9\linewidth]{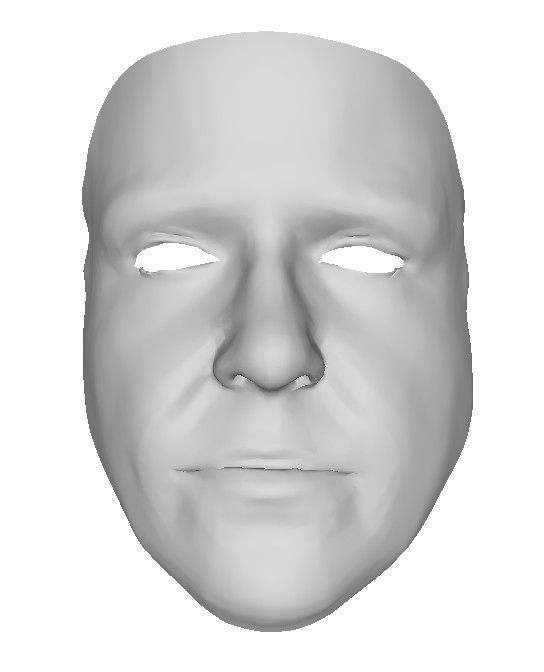}
			\subcaption{}
			\label{fig:shape-completion-b}
		\end{subfigure}
		\centering
		\begin{subfigure}{0.12\linewidth}
			\includegraphics[width=0.9\linewidth]{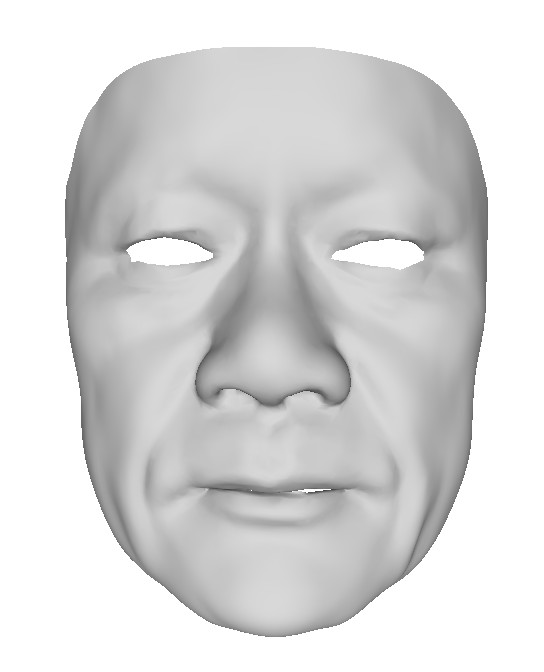}
			\subcaption{}
			\label{fig:shape-completion-c}
		\end{subfigure}
		\centering
		\begin{subfigure}{0.12\linewidth}
			\includegraphics[width=0.9\linewidth]{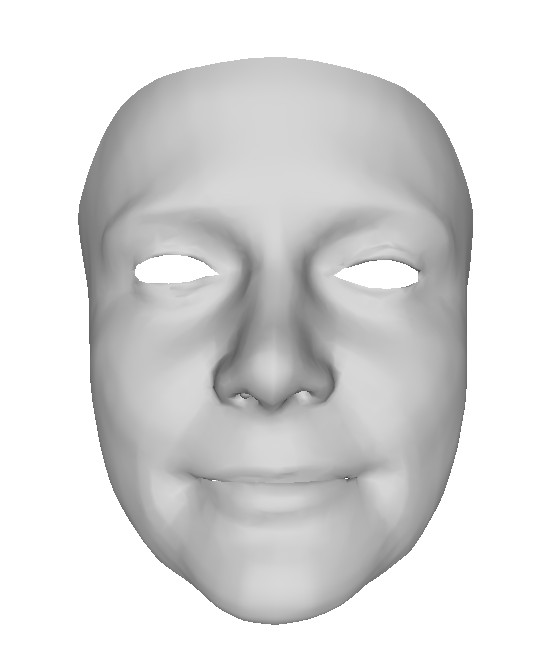}
			\subcaption{}
			\label{fig:shape-completion-d}
		\end{subfigure}
		\centering
		\begin{subfigure}{0.12\linewidth}
			\includegraphics[width=0.9\linewidth]{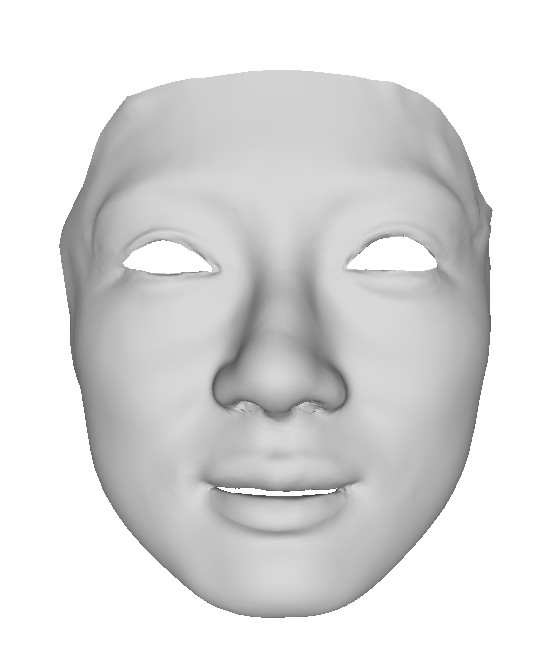}
			\subcaption{}
			\label{fig:shape-completion-e}
		\end{subfigure}
		\centering
		\begin{subfigure}{0.12\linewidth}
			\includegraphics[width=0.9\linewidth]{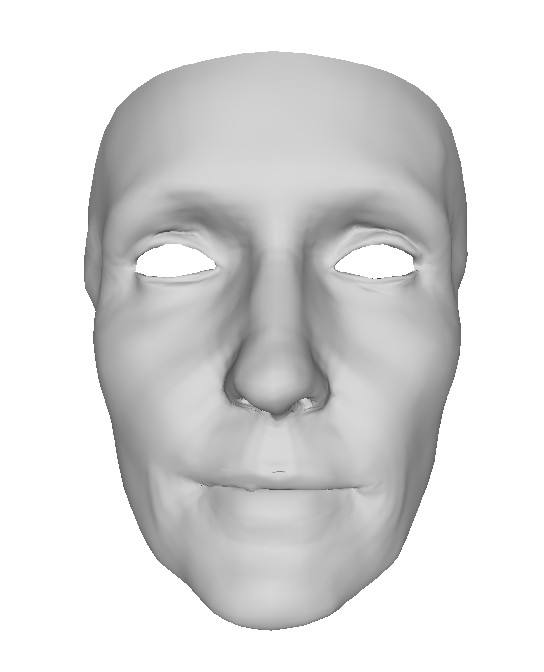}
			\subcaption{}
			\label{fig:shape-completion-f}
		\end{subfigure}
		\centering
		\begin{subfigure}{0.12\linewidth}
			\includegraphics[width=0.9\linewidth]{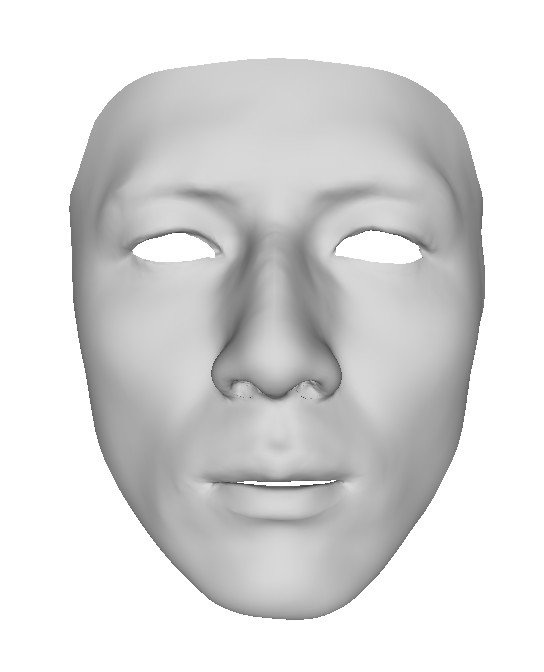}
			\subcaption{}
			\label{fig:shape-completion-g}
		\end{subfigure}
		\centering
		\begin{subfigure}{0.12\linewidth}
			\includegraphics[width=0.9\linewidth]{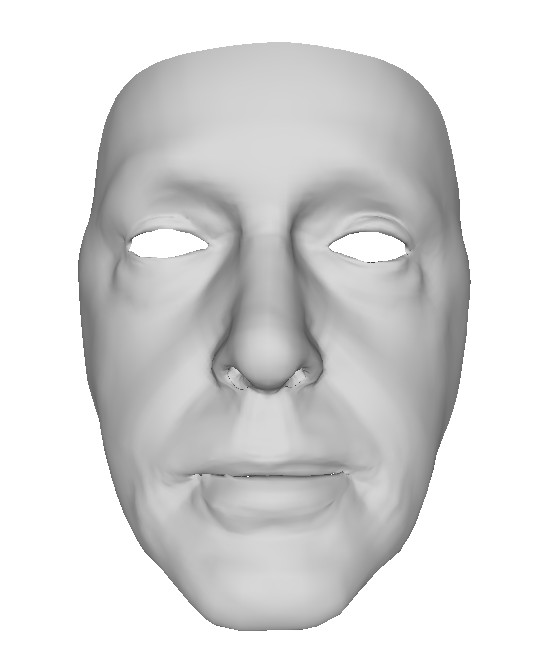}
			\subcaption{}
			\label{fig:shape-completion-h}
		\end{subfigure}
	\end{minipage}

	\vspace{5pt}
	\rule{\textwidth}{0.4pt}
	\vspace{-8pt}

	% 9-16
	
	\centering
	\begin{minipage}{0.02\linewidth}
		\rotatebox{90}{\textbf{defect}}
	\end{minipage}
	\begin{minipage}{0.93\linewidth}
		\centering
		\begin{subfigure}{0.12\linewidth}
			\includegraphics[width=0.9\linewidth]{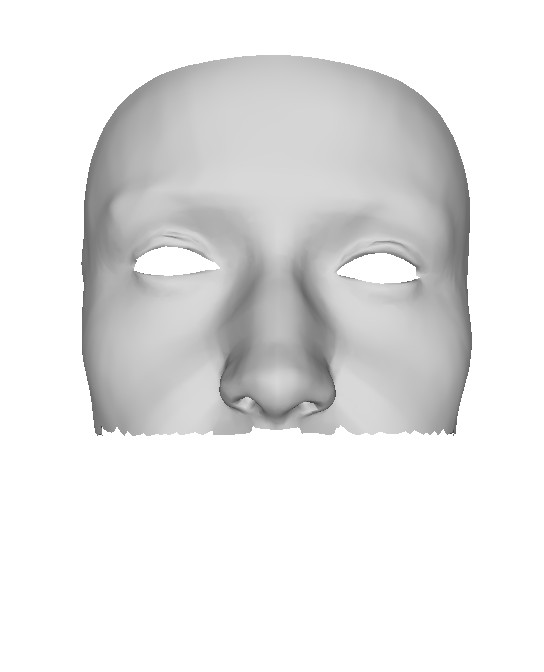}
		\end{subfigure}
		\centering
		\begin{subfigure}{0.12\linewidth}
			\includegraphics[width=0.9\linewidth]{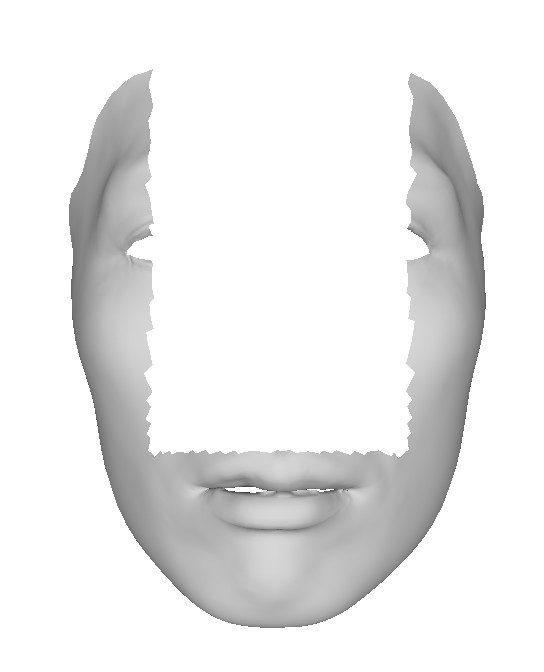}
		\end{subfigure}
		\centering
		\begin{subfigure}{0.12\linewidth}
			\includegraphics[width=0.9\linewidth]{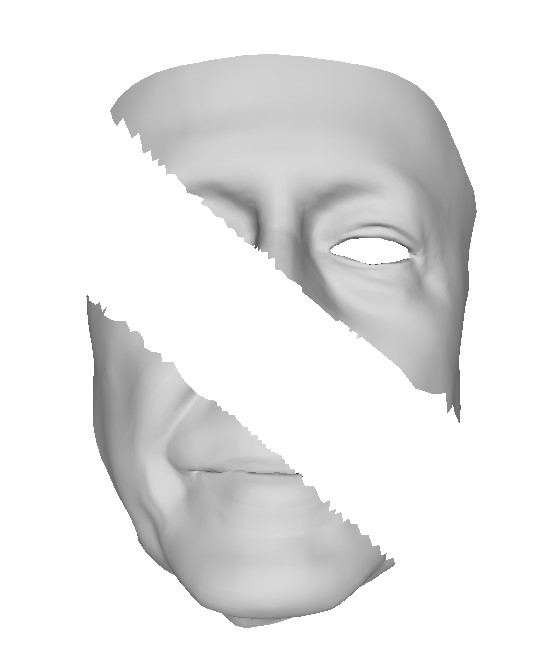}
		\end{subfigure}
		\centering
		\begin{subfigure}{0.12\linewidth}
			\includegraphics[width=0.9\linewidth]{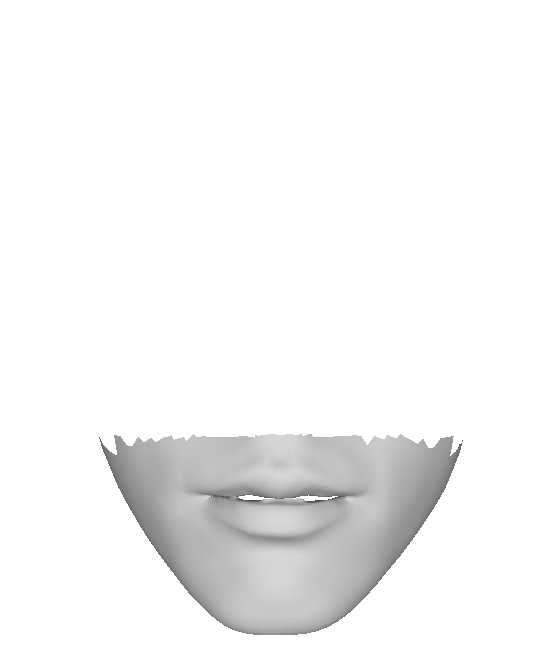}
		\end{subfigure}
		\centering
		\begin{subfigure}{0.12\linewidth}
			\includegraphics[width=0.9\linewidth]{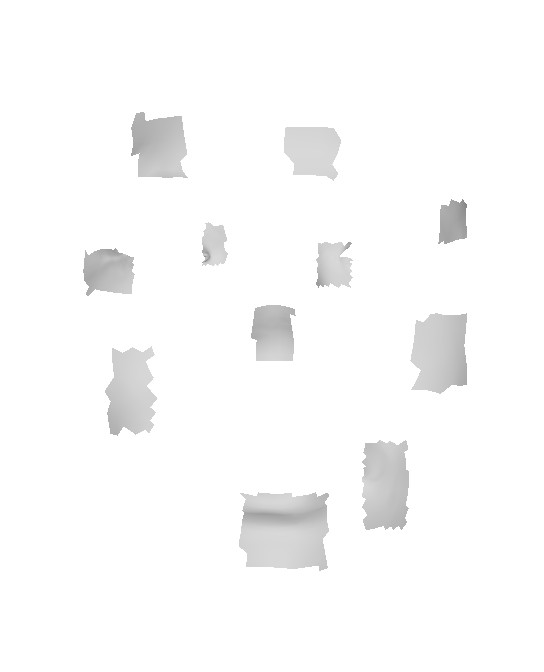}
		\end{subfigure}
		\centering
		\begin{subfigure}{0.12\linewidth}
			\includegraphics[width=0.9\linewidth]{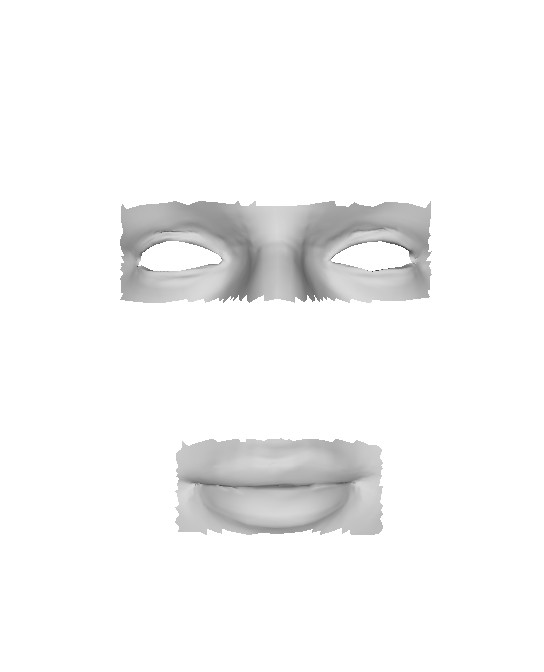}
		\end{subfigure}
		\centering
		\begin{subfigure}{0.12\linewidth}
			\includegraphics[width=0.9\linewidth]{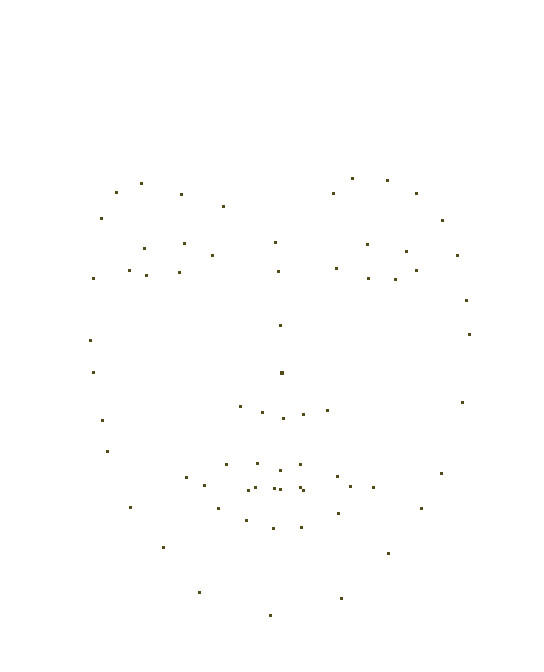}
		\end{subfigure}
		\centering
		\begin{subfigure}{0.12\linewidth}
			\includegraphics[width=0.9\linewidth]{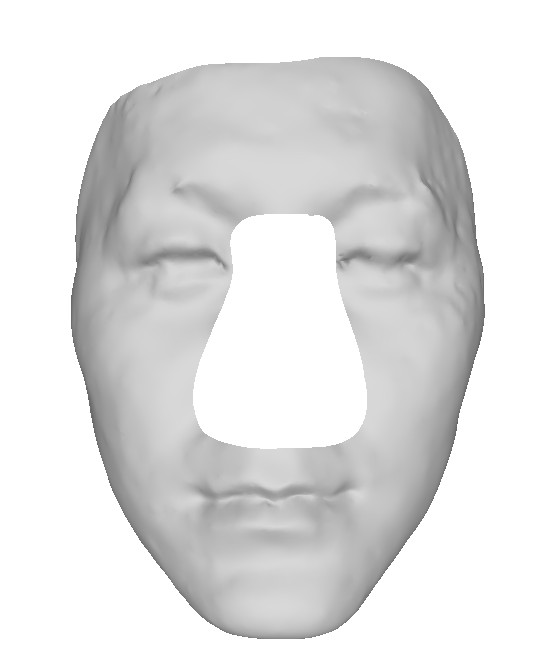}
		\end{subfigure}
	\end{minipage}
	
	\centering
	\begin{minipage}{0.02\linewidth}
		\rotatebox{90}{\textbf{result}}
	\end{minipage}
	\begin{minipage}{0.93\linewidth}
		\centering
		\begin{subfigure}{0.12\linewidth}
			\includegraphics[width=0.9\linewidth]{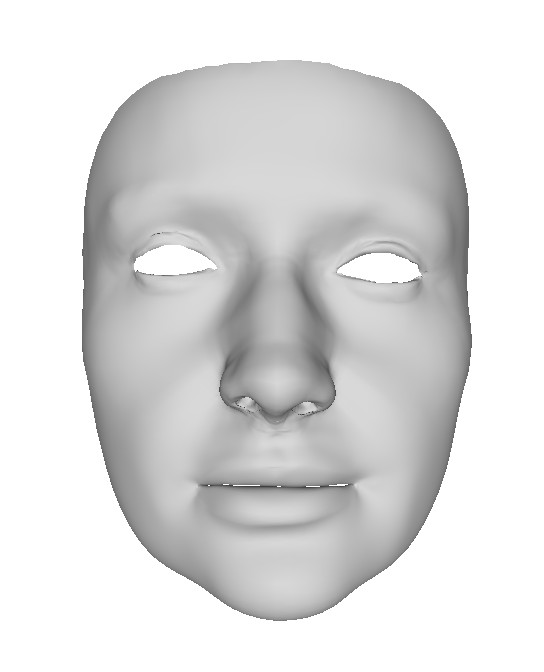}
		\end{subfigure}
		\centering
		\begin{subfigure}{0.12\linewidth}
			\includegraphics[width=0.9\linewidth]{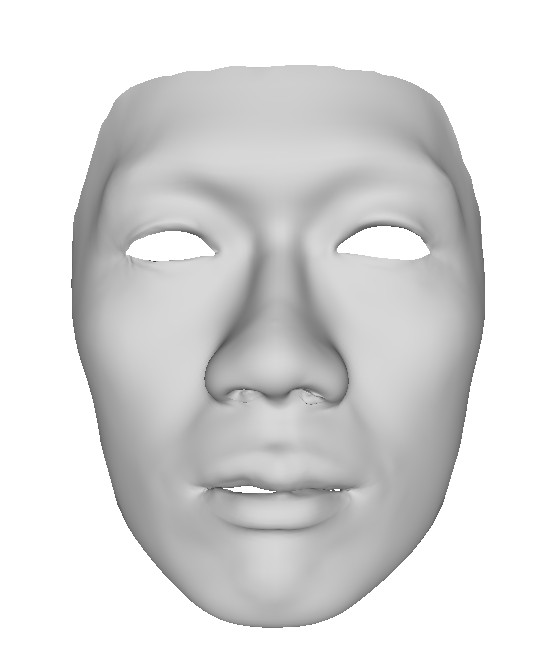}
		\end{subfigure}
		\centering
		\begin{subfigure}{0.12\linewidth}
			\includegraphics[width=0.9\linewidth]{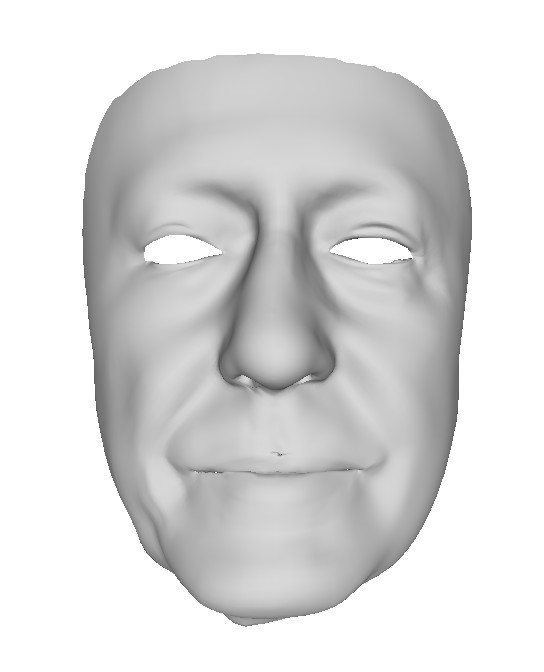}
		\end{subfigure}
		\centering
		\begin{subfigure}{0.12\linewidth}
			\includegraphics[width=0.9\linewidth]{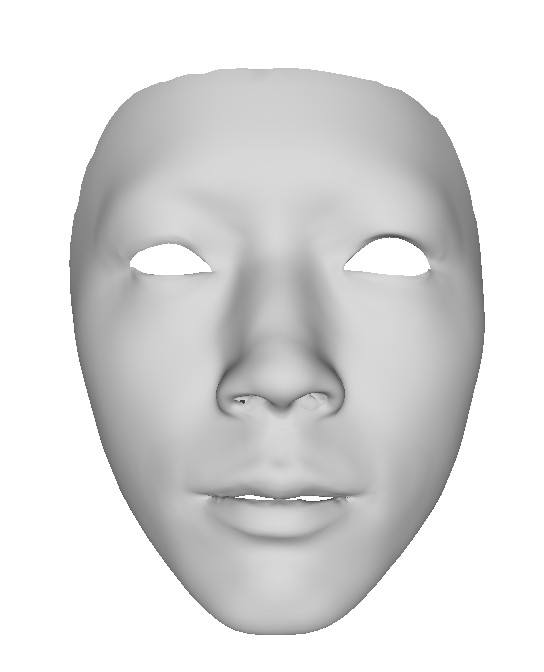}
		\end{subfigure}
		\centering
		\begin{subfigure}{0.12\linewidth}
			\includegraphics[width=0.9\linewidth]{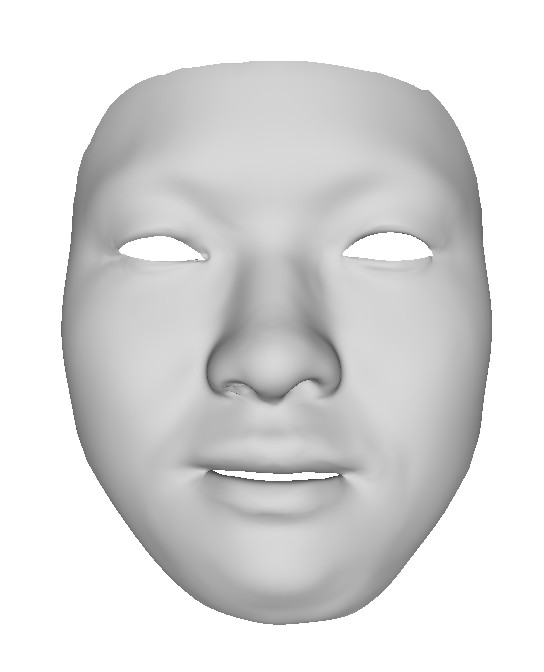}
		\end{subfigure}
		\centering
		\begin{subfigure}{0.12\linewidth}
			\includegraphics[width=0.9\linewidth]{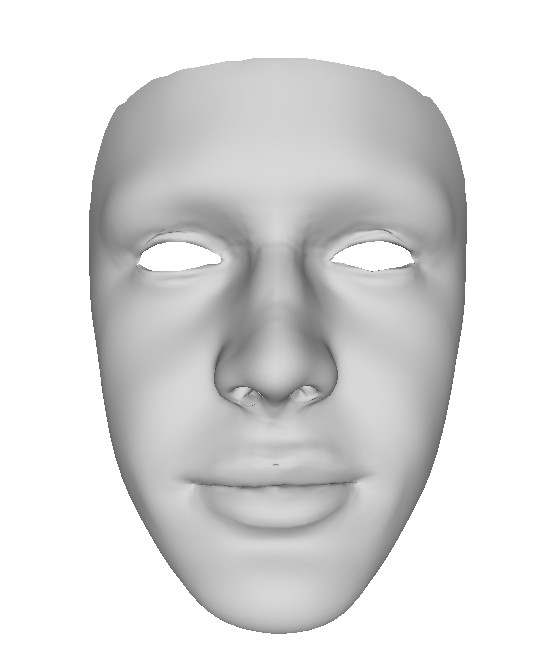}
		\end{subfigure}
		\centering
		\begin{subfigure}{0.12\linewidth}
			\includegraphics[width=0.9\linewidth]{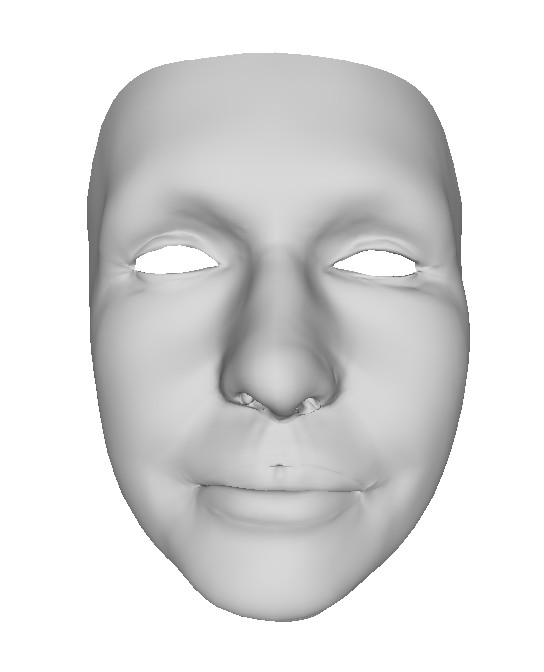}
		\end{subfigure}
		\centering
		\begin{subfigure}{0.12\linewidth}
			\includegraphics[width=0.9\linewidth]{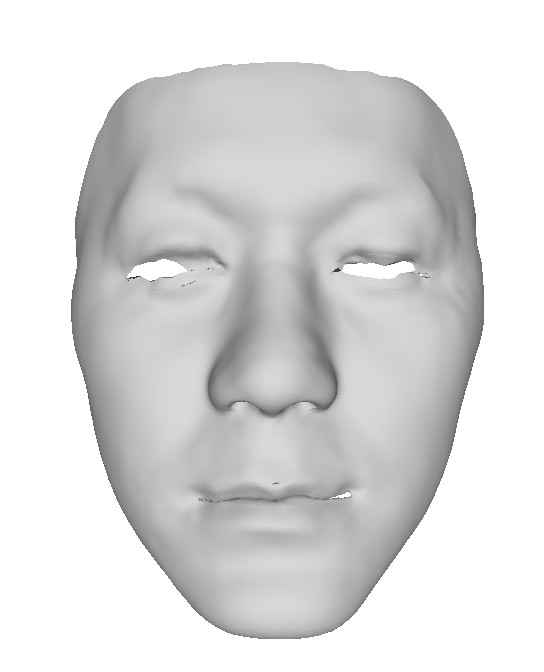}
		\end{subfigure}
	\end{minipage}
	
	\centering
	\begin{minipage}{0.02\linewidth}
		\rotatebox{90}{\textbf{gt}}
	\end{minipage}
	\begin{minipage}{0.93\linewidth}
		\centering
		\begin{subfigure}{0.12\linewidth}
			\includegraphics[width=0.9\linewidth]{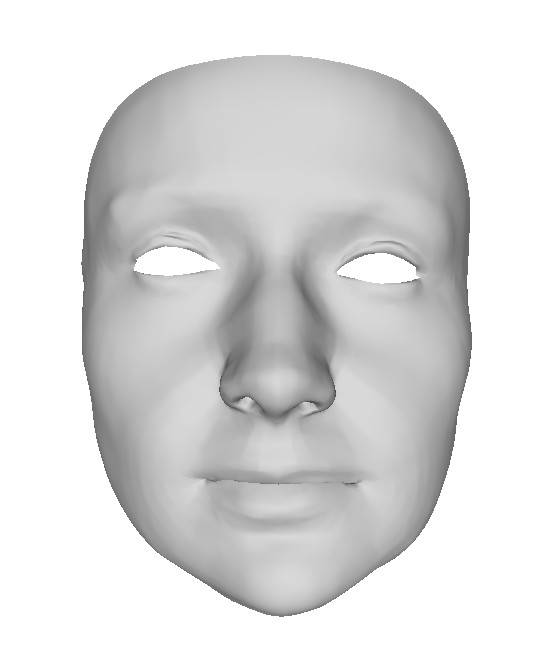}
			\subcaption{}
			\label{fig:shape-completion-i}
		\end{subfigure}
		\centering
		\begin{subfigure}{0.12\linewidth}
			\includegraphics[width=0.9\linewidth]{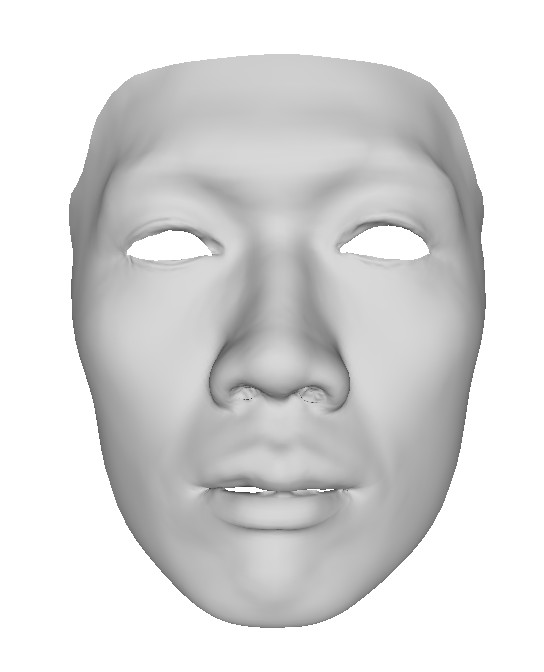}
			\subcaption{}
			\label{fig:shape-completion-j}
		\end{subfigure}
		\centering
		\begin{subfigure}{0.12\linewidth}
			\includegraphics[width=0.9\linewidth]{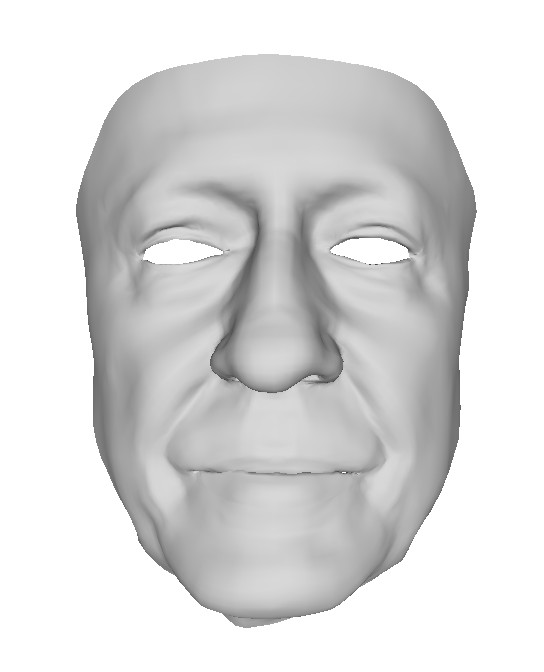}
			\subcaption{}
			\label{fig:shape-completion-k}
		\end{subfigure}
		\centering
		\begin{subfigure}{0.12\linewidth}
			\includegraphics[width=0.9\linewidth]{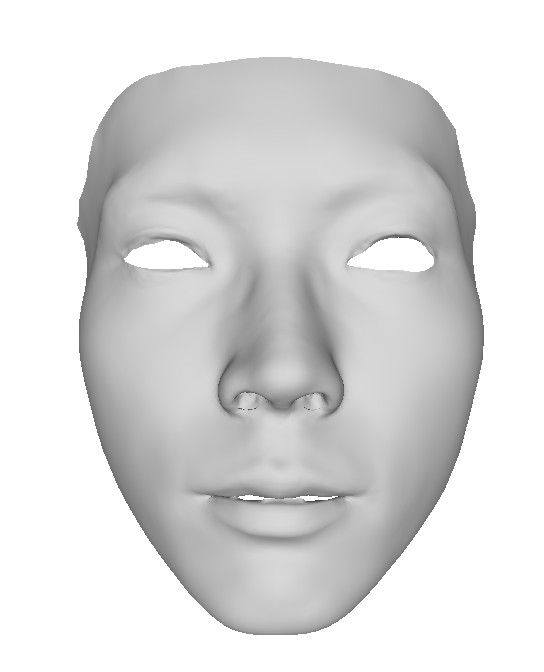}
			\subcaption{}
			\label{fig:shape-completion-l}
		\end{subfigure}
		\centering
		\begin{subfigure}{0.12\linewidth}
			\includegraphics[width=0.9\linewidth]{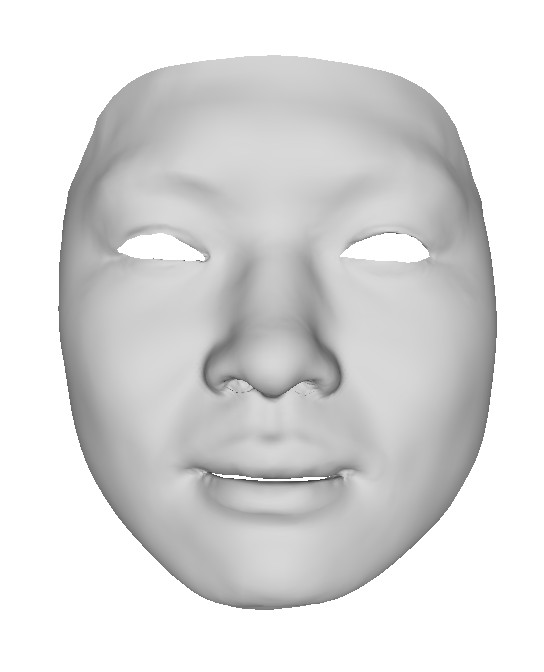}
			\subcaption{}
			\label{fig:shape-completion-m}
		\end{subfigure}
		\centering
		\begin{subfigure}{0.12\linewidth}
			\includegraphics[width=0.9\linewidth]{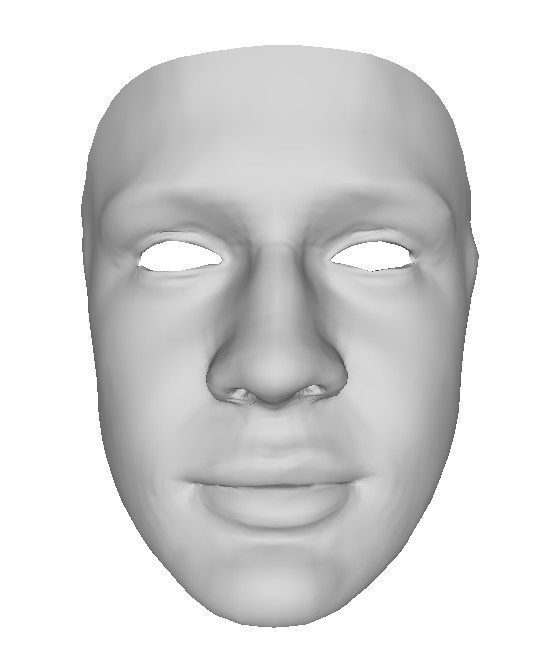}
			\subcaption{}
			\label{fig:shape-completion-n}
		\end{subfigure}
		\centering
		\begin{subfigure}{0.12\linewidth}
			\includegraphics[width=0.9\linewidth]{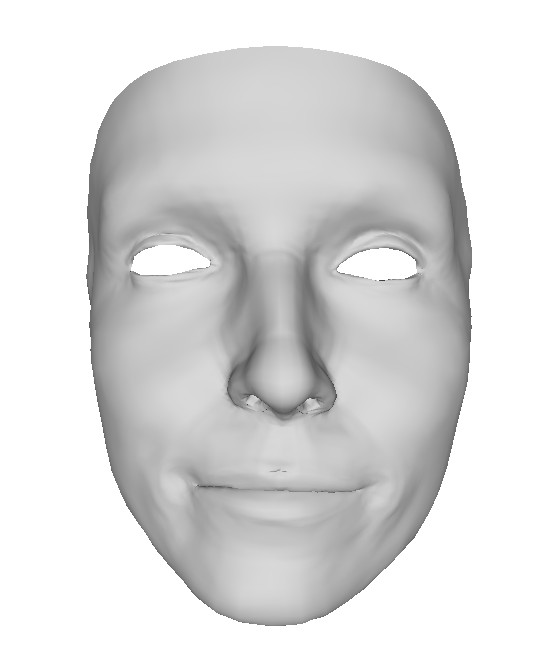}
			\subcaption{}
			\label{fig:shape-completion-o}
		\end{subfigure}
		\centering
		\begin{subfigure}{0.12\linewidth}
			\includegraphics[width=0.9\linewidth]{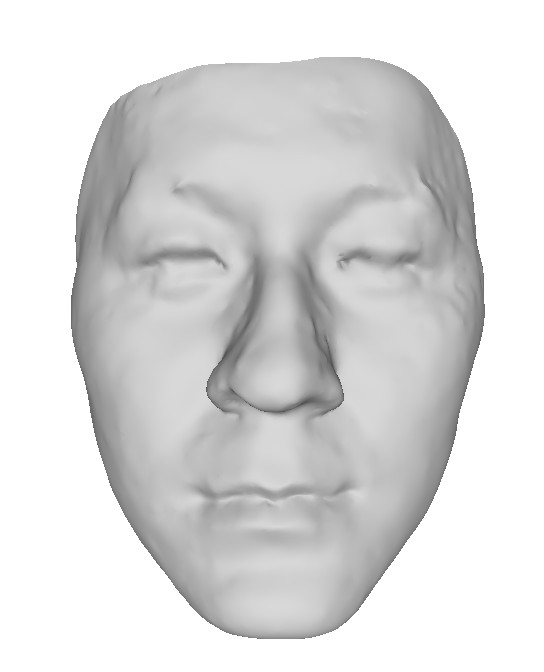}
			\subcaption{}
			\label{fig:shape-completion-p}
		\end{subfigure}
	\end{minipage}
	
	\caption{\textbf{Examples of facial shape completion experiment}. FaceCom demonstrates high-quality completion results for facial inputs with defects of various positions and sizes, even for samples with numerous wrinkles or lower quality. Please refer to our supplementary materials for more detailed views.}
	\label{fig:shape-completion}
\end{figure*}

To the best of our knowledge, there have been few works specifically addressing facial shape completion in the field of computer vision. The experiments of point cloud completion methods for facial datasets have yielded unsatisfactory results. Please refer to our supplementary materials for details. Therefore, in this section, we did not conduct comparative experiments. Instead, we tested FaceCom's completion capability by constructing several incomplete facial inputs on the test set with various regions and extents, and presented visual results.

The results of facial shape completion experiments are presented in \cref{fig:shape-completion}. The visualizations demonstrate the efficacy of FaceCom in achieving natural and high-fidelity completion results for defects located in different facial areas (\cref{fig:shape-completion-a,fig:shape-completion-b,fig:shape-completion-c,fig:shape-completion-d,fig:shape-completion-e,fig:shape-completion-f}), extensive defect areas (\cref{fig:shape-completion-g,fig:shape-completion-h,fig:shape-completion-i,fig:shape-completion-j,fig:shape-completion-k}), and cases involving elderly subjects with pronounced wrinkles (\cref{fig:shape-completion-c,fig:shape-completion-k}). Even with extremely limited inputs, such as partial information about specific regions (\cref{fig:shape-completion-l,fig:shape-completion-n}), scattered fragments (\cref{fig:shape-completion-m}), or just key points (\cref{fig:shape-completion-o}), FaceCom is capable of generating natural and authentic completion results. As the size of the defect increases, the information provided decreases, resulting in larger discrepancies between the completion results and the ground truth. For example, in \cref{fig:shape-completion-l,fig:shape-completion-n}, it is challenging to determine the original facial fullness based on the available information, leading to completion results that do not entirely match the actual condition. However, the non-defective regions remain consistent, resulting in a harmonious completion. It is important to note that our optimization-based completion strategy focuses on generating a complete facial structure rather than directly generating defect regions. While this approach proves effective for a wide range of irregular defect inputs and yields excellent fitting results, there might be slight mismatches in non-defective areas. We attempted to address this concern by employing post-processing techniques.

The defect test data we manually constructed were aimed at highlighting the capability of our method to effectively complete various irregular defect inputs. It should be noted that some of the defect styles, such as the one illustrated in \cref{fig:shape-completion-k}, are not typically encountered in clinical cases. To evaluate the practical application of our solution in a clinical setting, we tasked medical professionals with collecting several samples of nasal defects that conform to clinical conditions. One of these examples is depicted in \cref{fig:shape-completion-p}. The experimental results and analysis are outlined in \cref{clinical-experiment}.

\subsection{Clinical Experiment}
\label{clinical-experiment}

\begin{figure}[t]
	\centering
	\includegraphics[width=0.8\linewidth]{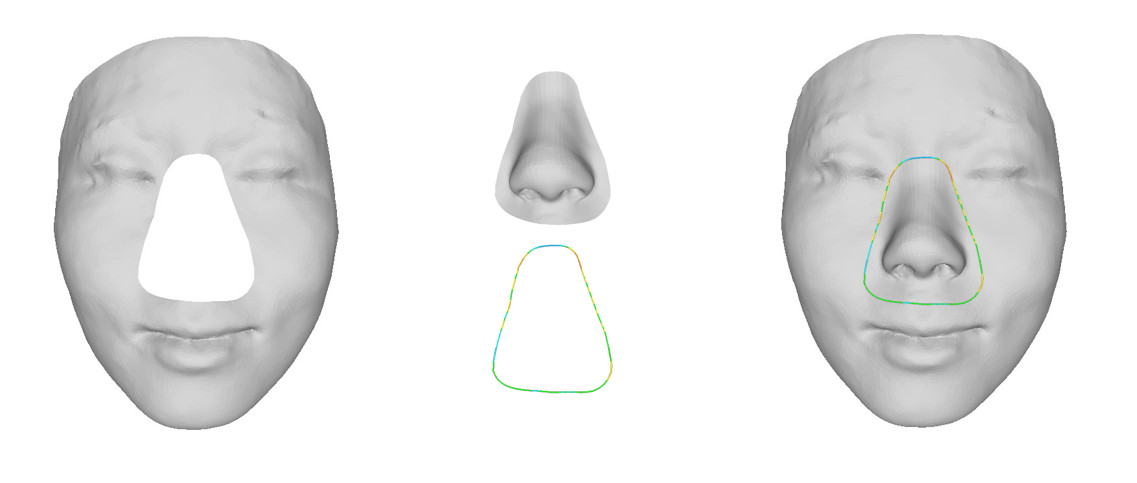}
	\caption{\textbf{Clinical Experiment}. Left: nasal defect input. Middle: extract nasal region from completion result. Right: combine and calculate margin fitness.}
	\label{fig:clinical}
\end{figure}

In clinical facial prosthetics, two main requirements are typically considered. Firstly, the prosthetic margins should seamlessly adhere to the contours of the patient's defect area. This criterion is evaluated by the root mean square distance of sample points, noted as margin fitness, and is deemed the most critical factor. Secondly, the prosthesis should blend naturally with the patient's original facial features, which is challenging to quantify and is often assessed visually.

To test the practical application of FaceCom in a clinical setting, we recruited 20 clinical patients in the hospital and scanned their facial data to construct nasal defects. These 20 cases of nasal defects were all handled by professional physicians and met the standards for clinical data collection. Following the completion process by FaceCom, we extracted the nasal region for evaluation, as illustrated in \cref{fig:clinical}. The average margin fitness of the completed results for the 20 cases of nasal defects was 0.33±0.21 mm, indicating their practicality for manufacturing and use. Additionally, the natural and seamless appearance of the completed results has been acknowledged by the physicians. Our method shows promise in providing assistance to facial defect patients in the medical field.

\subsection{Non-rigid Registration}
\label{non-rigid-registration}

While our method is primarily designed for completing defective 3D facial scans, it can also be utilized as a non-rigid face registration tool for facial scans with potential holes or missing areas. This is because FaceCom maintains topological consistency of outputs across various facial scans. To assess FaceCom's performance in non-rigid facial registration, we conducted a simple test. Using the same test data as in \cref{fitting}, we applied the full completion approach to achieve non-rigid registration results. The quantitative results of the registration are shown in the last row of \cref{tab:fit-compare}. Although we did not conduct a comparative experiment for non-rigid registration, as it falls outside the primary focus of this paper, our method achieves a precision of 0.16mm for the mean surface error, similar to the facial non-rigid registration algorithm proposed in \cite{fan2023towards}, without the need for landmarks. This underlines the potential of our method in the field of non-rigid facial registration.

\subsection{Ablation Study}
\label{ablation-study}

\begin{table}
	\centering
	\begin{tabular}{lcc}
		\toprule
		Method & reconstruction & regularization \\
		\midrule
		Ours(global-only) & 0.336 & \textbf{0.240} \\
		Ours(local-only) & 0.208 & 0.419 \\
		Ours(full) & \textbf{0.153} & 0.419 \\
		\bottomrule
	\end{tabular}
	\caption{\textbf{Ablation experiment results for shape generator}.}
	\label{tab:ablation}
\end{table}

\begin{figure}[t]
	\centering
	\begin{subfigure}{0.22\linewidth}
		\includegraphics[width=0.99\linewidth]{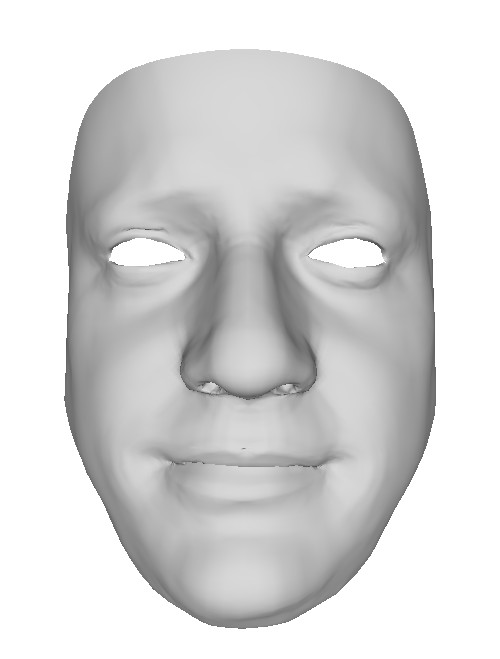}
		\caption{}
		\label{fig:ablation-a}
	\end{subfigure}
	\centering
	\begin{subfigure}{0.22\linewidth}
		\includegraphics[width=0.99\linewidth]{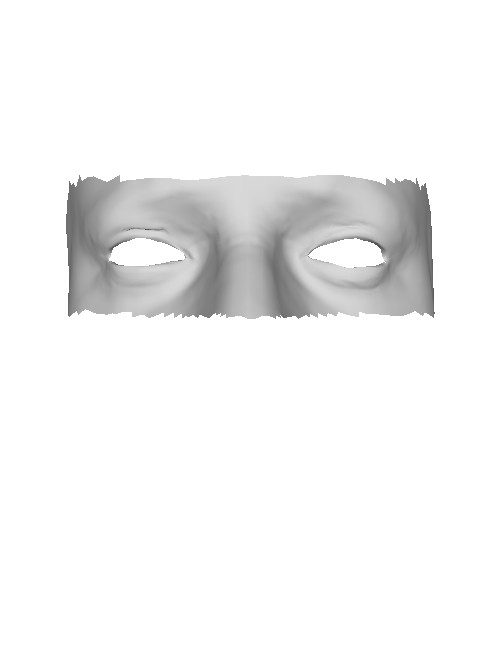}
		\caption{}
		\label{fig:ablation-b}
	\end{subfigure}
	\centering
	\begin{subfigure}{0.22\linewidth}
		\includegraphics[width=0.99\linewidth]{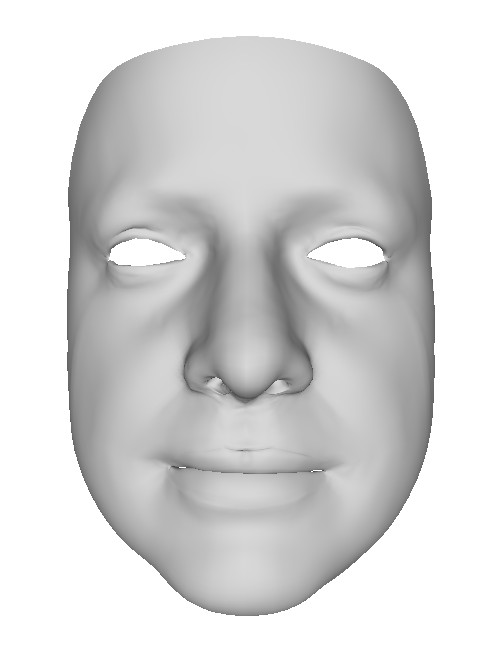}
		\caption{}
		\label{fig:ablation-c}
	\end{subfigure}
	\centering
	\begin{subfigure}{0.22\linewidth}
		\includegraphics[width=0.99\linewidth]{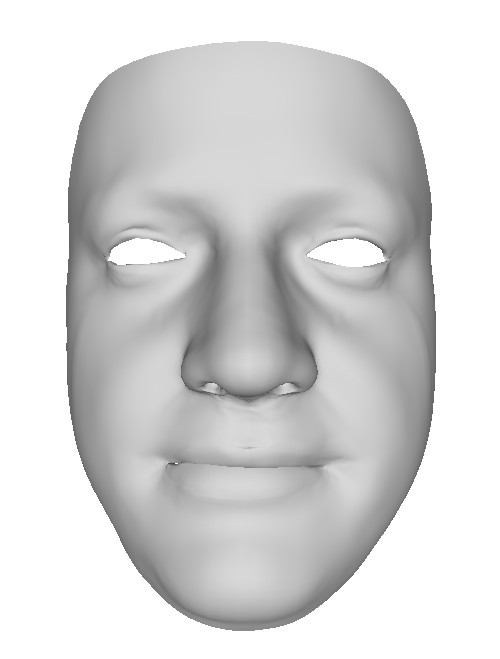}
		\caption{}
		\label{fig:ablation-d}
	\end{subfigure}
	
	\caption{\textbf{Ablation experiment example for image guidance}. \subref{fig:ablation-a} Ground truth. \subref{fig:ablation-b} Constructed large-scale defect. \subref{fig:ablation-c} Fitting result without image guidance. \subref{fig:ablation-d} Fitting result with image guidance. }
	\label{fig:ablation}
\end{figure}

In this section, we conduct ablation experiments on the shape generator to assess the significance of its local and global modules. Additionally, we examine the role of image inpainting guidance on the 3D facial shape completion process.

For the shape generator, we isolate the global and local modules individually while adjusting the channel numbers to maintain a consistent parameter count. We trained each of them under the same training parameters. The experimental results, including the quantitative reconstruction loss and regularization loss, are summarized in \cref{tab:ablation}. Notably, we observe that under similar training conditions, the network with only the global module tends to prioritize optimizing the regularization term, whereas the full network demonstrates superior mesh data reconstruction.

For the image inpainting guidance module, we explore its influence on FaceCom's completion results and provide qualitative analysis. Our findings suggest that for cases with smaller defect areas, the image inpainting guidance has minimal impact. Conversely, for scan data with more extensive defect areas, as illustrated in \cref{fig:ablation}, this method achieves results that are closer to the ground truth, effectively steering the shape generator's generation process.

\section{Discussion}
\label{discussion}

Due to the intended application of our research in providing assistance for patients with facial defects, we primarily considered only basic shape variations. However, this approach may limit its applicability in other contexts. For future improvements, incorporating changes such as expressions and textures into our facial shape completion solution could be explored. Additionally, addressing the issue of less effective deformation handling at the edges during geometric processing is another aspect that requires attention.

The publicly available datasets used in this research were obtained through formal applications and adhered to their respective agreements. Additionally, the clinical facial data we collected was obtained with the consent of the patients and has been approved for scientific research by the relevant Institutional Review Board. We have provided the corresponding approval documents in the supplementary materials.

\section{Conclusion}
\label{conclusion}

We propose a facial shape completion solution FaceCom in this paper, involving training a shape generator on diverse individual datasets, followed by the generation of a complete face from the incomplete input using optimization and image inpainting guidance, and finally refining the completed face through post-processing. The entire process is automated and efficient, yielding high-fidelity completion results for irregular incomplete facial scans. Our approach demonstrates promising results in fitting and completion, and it can be practically applied to the design of clinical facial prostheses. 

\noindent\textbf{Acknowledgment}. The authors would like to thank the anonymous reviewers for their suggestions in improving this paper. They would also like to thank Junlin Chang for his insightful ideas. This work is supported in part by the National Natural Science Foundation of China under Grant 82271039 and 62132021, in part by the Open Project Program of Peking University School of Stomatology under Grant PKUSS20220202.

%%%%%%%%% REFERENCES
{\small
	\bibliographystyle{ieee_fullname}
	\bibliography{ref}
}

\end{document}